\documentclass[preprint,3p,12pt]{elsarticle}
\usepackage{tcolorbox}
\def\thedemobiblio#1{\smallskip\par
 \list{}{\labelwidth 0pt \leftmargin 1em \itemindent -1em \itemsep 1pt}
 \small \parindent 0pt
 \parskip 1.5pt plus .1pt\relax
 \def\newblock{\hskip .11em plus .33em minus .07em}
 \sloppy\clubpenalty4000\widowpenalty4000
 \sfcode`\.=1000\relax}

\newcommand{\real}{\hbox{\rm I\kern-0.2emR}}

\usepackage{amssymb}
\usepackage{amsmath}
\usepackage{amsfonts}
\usepackage{amsbsy}
\usepackage{listings}

\usepackage{geometry}
\usepackage{acronym}
\usepackage{hyperref}

\usepackage{color}
\usepackage{float}
\usepackage{graphicx}
\usepackage{tcolorbox}
\usepackage{caption}
\usepackage{subcaption}
\usepackage{lineno}
\usepackage{epstopdf}
\usepackage[utf8]{inputenc}
\usepackage{fancyhdr}
\usepackage{multirow}
\usepackage{tikz}
\usetikzlibrary{trees}
\usepackage{rotating}
\usepackage{ulem}
\usepackage{algorithm}
\usepackage{algpseudocode}
\usepackage{algcompatible}

\journal{}

\begin{document}

%\linenumbers

\title{\large {\bf
A Real-time Multimodal Transformer Neural Network-powered Wildfire Forecasting System
}
}

\author{Qijun Chen ${}^{\dag}$}
\author{Shaofan Li ${}^{\dag}$\footnote{Email:shaofan@berkeley.edu}}

\address{
$^{\dag}$ Department of Civil and Environmental Engineering,
University of California, Berkeley, \\
California, 94720, USA; \\
}


\abstract{
Due to climate change,
extreme wildfire has become one of the most dangerous natural hazards to human civilization.
Even though, some wildfires may be initially caused by human activity, the spread
of wildfires is mainly determined by environmental factors, for example,
(1) weather conditions such as temperature,
wind direction and intensity, and moisture levels;
(2) the amount and types of dry vegetation in a local area,
and (3) topographic or local terrain conditions, which affect how much rain an area gets
and how fire dynamics will be constrained or facilitated.
Thus, to accurately forecast wildfire occurrence
has become one of the most urgent and taunting environmental challenges on a global scale.
In this work, we developed a real-time
Multimodal Transformer Neural Network Machine Learning model
that combines several advanced artificial intelligence techniques
and statistical methods
to practically forecast the occurrence of wildfire at the precise location
in real-time, which not only utilizes
large-scale data information such as hourly weather forecasting data,
but also takes into account small-scale topographical data
such as local terrain condition and local vegetation conditions
collecting from Google Earth images
to determine the probabilities of wildfire occurrence location
at small scale
as well as their timing synchronized with weather forecast information.
By using the wildfire data in the United States from 1992 to 2015 to train
the multimodal transformer neural network,
it can predict the probabilities of wildfire occurrence
according to the real-time weather forecast and
the synchronized Google Earth image data to provide
the wildfire occurrence probability
in any small location ($100m^2$)
within 24 hours ahead.
}\\
\smallskip
\smallskip

{\bf Keywords: Multimodal transformer neural network, Machine learning,
Extreme wildfire, Climate change, Convolutional
neural network}

}
%%


\maketitle

\section{Introduction}

Due to climate change, extreme wildfire has become one
of the most frequently happening, widely spreading,
and most dangerous hazards on Earth.
For example,
the direct property damage from the 2025
wildfire in Los Angeles County
is estimated at between \$28 billion and \$53.8 billion,
according to a report commissioned by the Southern California Leadership Council.
Nearly \$150 billion loss
in direct damages has been caused by California's 2018 wildfire alone.
In addition, wildfire smoke also causes significant environmental pollution.
According to an estimate from a report \cite{Samborska2024},
Global forest fires emitted  5 to 8 billion tonnes of $CO_2$ each year,
and it is more than 10\% of the global energy-related $CO_2$ emission.

\begin{figure}[!htb]
\begin{minipage}{0.53\linewidth}
    \begin{center}
    \includegraphics[width=3.3in]{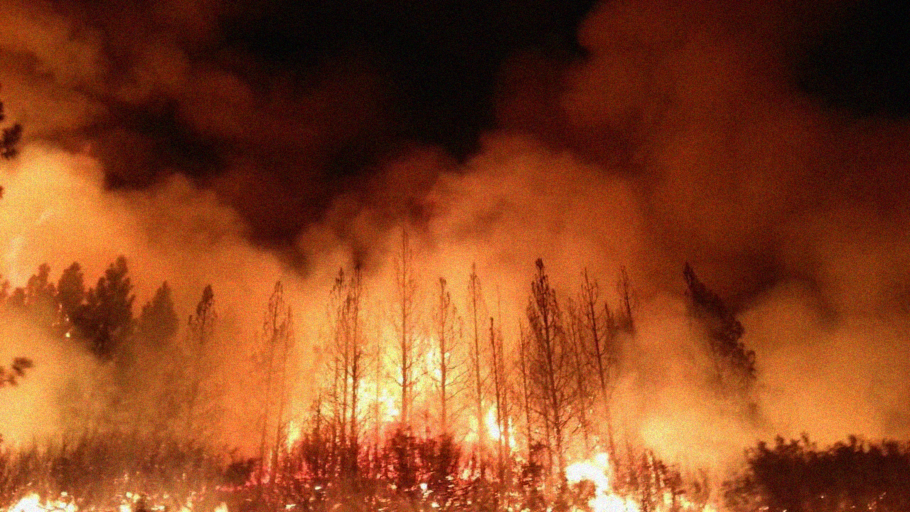}
     \end{center}
    \begin{center}
    (a)
    \end{center}
\end{minipage}
\begin{minipage}{0.44\linewidth}
    \begin{center}
    \includegraphics[height=1.9in]{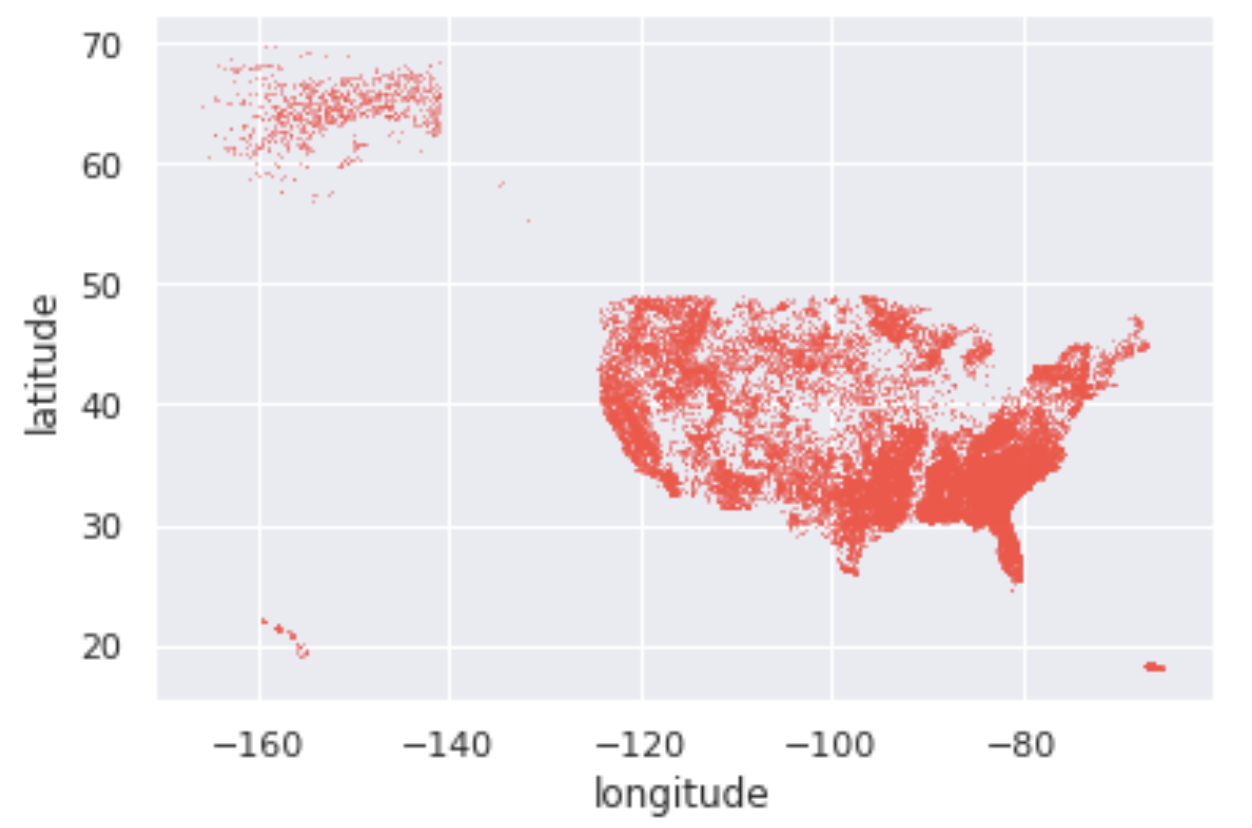}
     \end{center}
    \begin{center}
    (b)
    \end{center}
\end{minipage}
    \caption{(a) 2018 California Camp Fire
    [Photo: U.S. Department of Agriculture/Wikimedia Commons], and (b)
    Geographical locations of wildfires occurred in the United States}
    \label{fig:firemap}
\end{figure}

In particular, the 2025 wildfire in Los Angeles County
have killed at least 29 people, and forced more than 200,000 people to evacuate.
While in the 2018 California Camp Fire (see Figure \ref{fig:firemap} (a)),
there were 85 confirmed deaths, 17 injuries, and more than $52,000$ people
were evacuated.
In that fire,
more than 18,804 building structures
were destroyed, and the burn area exceeded $621~ km^2$.
The Camp Fire caused
more than 16.65 billion dollars in direct damage alone (2018 USD)).
Figure \ref{fig:firemap} (b) shows the geological distribution of
the wildfire that occurred in the United States.

By accurately forecasting the wildfire occurrence location and its timing,
we cannot only reduce the indiscriminate power shutdown in the fire season,
which causes power outages resulting of closing businesses and schools,
but also allow us to take active specific and direct
preventive measures in surgical precision.
For instance, if we know the precise fine-scale fire occurrence location,
we can control the vegetation growth and state in the specific
predicted location in a relatively short time with low cost and high efficiency.

Since the 2010s, several studies have used
artificial intelligence-related or machine learning (ML) based methods
to predict wildfire occurrences in a relatively large area e.g. \cite{Bolt2022,Dabrowski2023}.
However, most of these research works are still at a scholarly research level
but not at the practical fire-hazard prevention stage, such as
forecast the precise occurrence location within $1$ mile
radial distance.

The current ML-based wildfire prediction models
used the existing wildfire data and large-scale weather data
to predict possible fire locations.
Even though the existing machine-learning neural network models
may be able to predict some fire occurrences, the predicted location
is inside a large area $(> 10 ~mile^2)$.
However,
the actual and precise wildfire occurrence location in a landscape
is strongly influenced by local weather, terrain, and vegetation conditions.
Precisely speaking, the precise
fire occurrence location depends on
the small-scale weather, terrain, and vegetation conditions
that cannot be obtained from the existing database or weather forecasting,
but can only be extrapolated or correlated with the local terrain and
vegetation conditions,
and some of them can only be represented by the spatial conditional probability distribution.

Moreover, in some advanced studies on machine learning-based wildfire occurrence predictions,
most researchers adopt the logistic regression statistical
analysis (correlation) and artificial neural network (ANN) approach to
directly predict fire occurrence by adding vegetation cover\cite{cavard2015},
slope, elevation, the density of livestock, precipitation conditions,
and lightning polarity as features
(See: \cite{kalabokidis2016,duncan2010,matt2021}).% malik12010109}.
However, the spatial scale of these features is either inconsistent with
the scale of weather forecasts
or incompatible with the length scale of the existing wildfire data.

Since wildfire occurrence and its severity in a landscape is strongly
influenced by both weather and terrain conditions \cite{taylor2021}
at different locations, and in different spatial and temporal scales,
By categorizing the different causes of wildfire, and using and learning from the vast wildfire data,
in this work, we are developing a multiscale multimodal learning neural
network model that can provide the conditional probability
of the wildfire for different causes,

Unlike the current straightforward and elementary machine
learning approach,
the proposed Machine Learning Neural Network approach
can offer an efficient tool to practically forecast
the probability of wildfire occurrence in a small-scale spatial location
and a precise time frame.
It not only takes into account large-scale regular weather and historic wildfire data
such as hourly weather forecasting data, but also utilizes small-scale geographic
information such as local terrain conditions and local vegetation conditions
extrapolated from Google Earth images
to obtain the probabilities of wildfire occurrence with respect
to small-scale location as well as their timing.

%%XX

\section{Data Collections and Data Pre-processing}
In this work, the wildfire data collected are only within
the United States to demonstrate the feasibility of the developed machine learning neural network.

\begin{figure}[!htb]
    \centering
    \includegraphics[width=0.5\columnwidth]{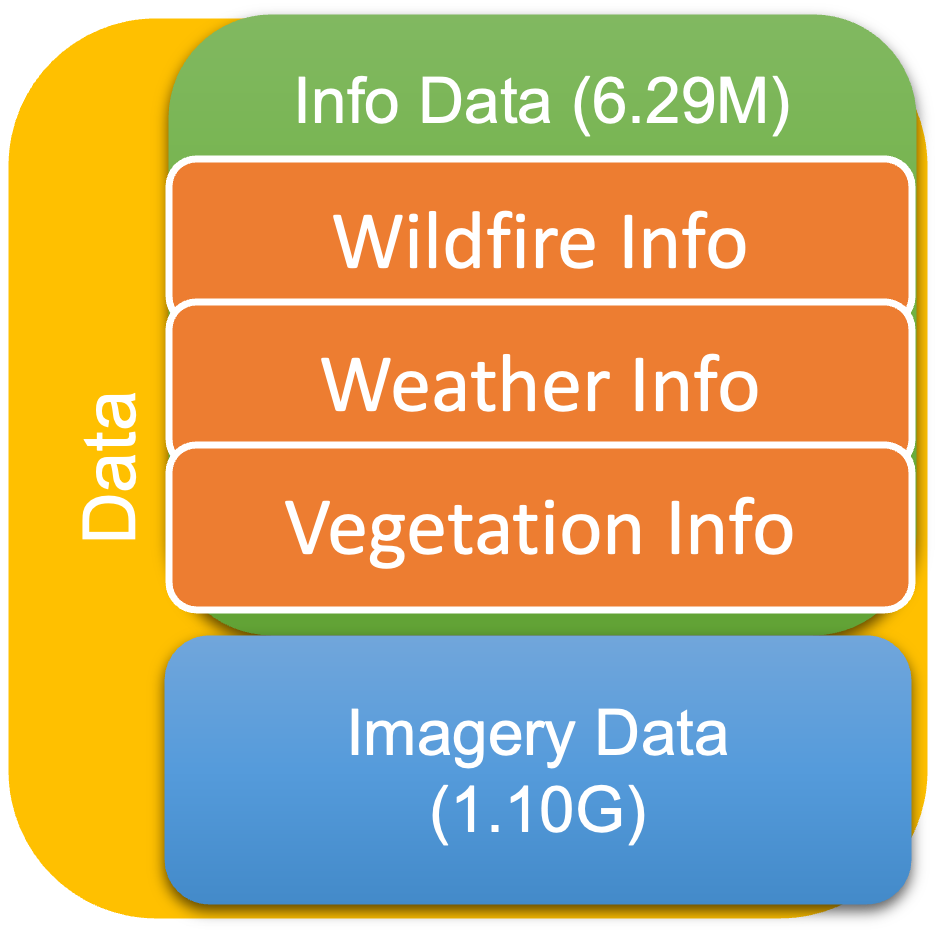}
    \caption{Data Components}
    \label{fig:data_components}
\end{figure}

As shown in Figure \ref{fig:data_components}, the wildfire data is mainly consists of two parts.
The first part is the information data, and the second part is the imagery data.
Within the first part, information data is a combination of three categories or sets of data:
wildfire information set, weather information set, and vegetation information set.
The first data set is wildfire information, which
consists of geo-referenced wildfire records
from the United States Department of Agriculture (USDA) \cite{fpa_fod2017}
from 1992 to 2015 of 24 years,

The geo-referenced records mainly include the fire location,
which is latitude and longitude, the size of the fire, which
is treated as quantitative data, and the causes of fire including
the date of the wildfire, which is also qualitative data.

In this study,
the qualitative data will be converted into quantitative
data by using methods such as One-Hot Encoding \cite{onehot}.
The miscellaneous data such as the fire name,
and the fire ID are not included in this class of data.
The second data set, weather information, contains mainly Integrated Surface Data (ISD).
It contains the records of the weather conditions
before the wildfire occurred.
For instance, it may contain the 7 days hourly average temperature
before the discovery date in the wildfire records.
The vegetation information will contain the categories of the location
that wildfire was occurred in the wildfire records.
The second part is the imagery set that is manually obtained by software
such as Google Earth Pro\cite{googleearthpro}.
It contains the obtainable images at the specific time when the fire
occurred within a certain radius from the fire locations
as the center with a clear view of the surface situations.

Since some data may not be available when combining two parts,
we remove the data with missing values.

\subsection{Information Data}

The information data refers
to the general information of a wildfire, and
it mainly consists of three parts:
wildfire records, the associated weather information, and the local
vegetation information.

In this work, the wildfire data records are obtained
from the Fire-Occurrence Database, FPA$\_$FOD, and
the data used in this work contains
spatial wildfire occurrence data for the United States from 1992-2015.
Since this database is kept renewed, currently, the FPA$\_$FOD database
has all the data collected from USDA from 1992-2022 \cite{fpa_fod2017}.
This wildfire database has its special data format,
such as Object\_ID, Source\_SYSTEM ID,  Report agency ID, LOCAL\_FIRE record ID,
etc. (see  Figure \ref{data_desc}).
When extrapolating the wildfire data from the database,
the essential information is maintained, which includes the latitude
of the fire occurrence location, the longitude of the fire location,
which is at least with the precision as required in Public Land Survey System
(PLSS) \cite{fpa_fod2017},
the size of the fire, the cause of the ire, date when the fire was discovered.

The next set of information data is the hourly weather data, which
is obtained from the National Centers
for Environmental Information
(NCEI) database that uses the nearest
NOAA station\cite{ncei} based on the latitude and longitude from previous
information databases.
The hourly weather data includes average temperature, wind speed, humidity,
and precipitation in $^{\circ}C$, $m/s$, $\%$, and $mm$ separately
at the location of the wildfire mentioned above up to 30, 15,
and 7 days before the discovery of the wildfire.
Since the recorded time from $FPA\_FOD$ is the discovery time,
getting the exact time is difficult,
in order to reduce the inaccuracy from the records,
the average information will be applied.
For example, we take the average of $avg(30 \times 24)$,
$avg(15 \times 24)$, and $avg(7 \times 24)$ and keep three numbers
as its temperature value.

\begin{figure}[!htb]
    \centering
    \includegraphics[width=3.5in]{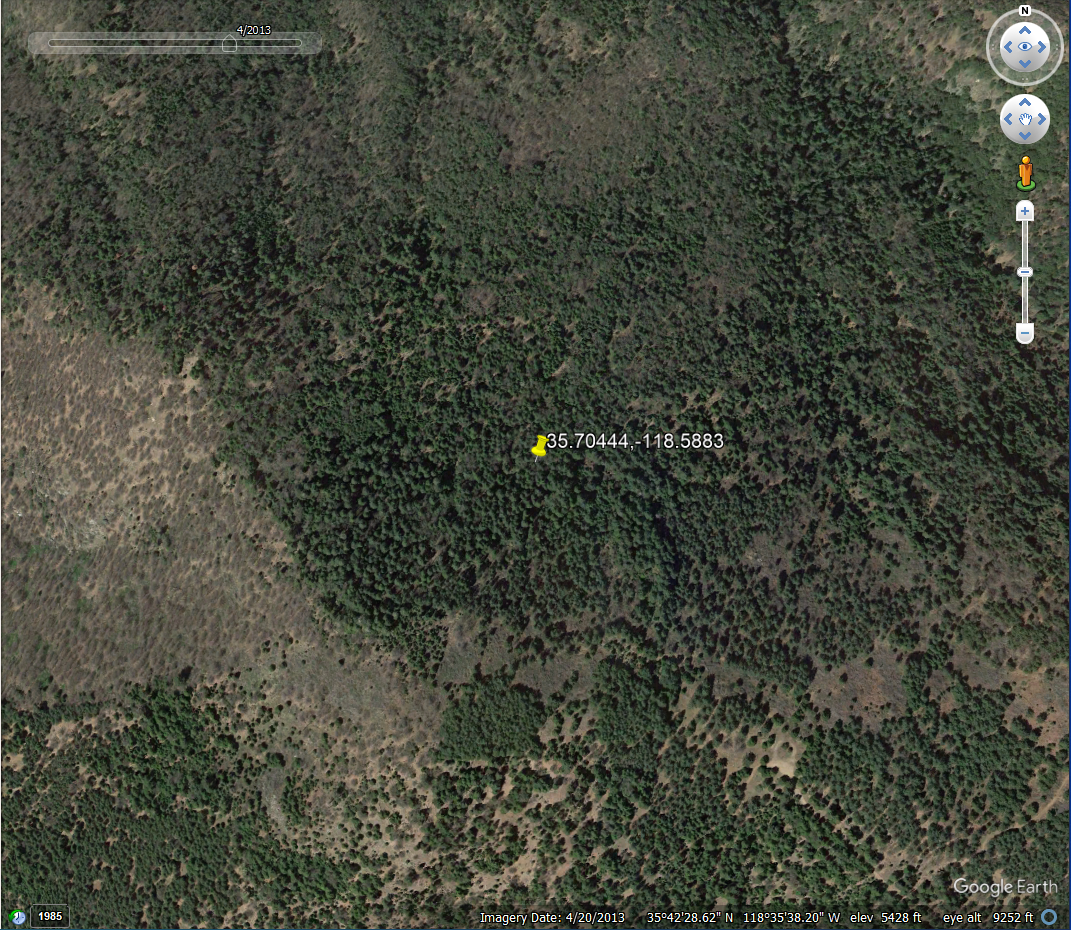}
    \caption{An example of image file from Google Map @35.7044,-118.5883333, on 8/18/2013}
    \label{fig:googlemapexample}
\end{figure}

The vegetation information contains the vegetation
categories of the location where the fire occurred.
It is available at OPeNDAP\cite{opendap} dataset,
a remote data retrieval software.
For example, for the case of fire,
as shown in a Google Map capture in Figure \ref{fig:googlemapexample},
caused by lightning occurred at location $(35.7044444N, -118.588333W)$ in date $8/18/2013$,
which is in Florida, and it has a fire size of $76$, and the average temperature, wind, humidity,
and precipitation data before the fire occurrence date in 7, 15, 30 days,
and the vegetation type is characterized as the shrubland type, or shrubland.

The original data from $FPA\_FOD$ for fire information has 1.88 million entries.
It is an enormous amount of data and not all of the entries can obtain enough
information such as weather or vegetation.
We also combined with a exist data that is available
from GitHub for the information set\cite{varun2020}.
Currently, with the combined part, the total amount of data obtainable is about $29550$.
Figure \ref{fig:firemap} shows the plot of latitude and longitude
from the current data for all fire cases and time.
The information data itself can also be used
to train the baseline model.
Since in this work the model's neural network architecture has been developed,
in the future, once the new data become available,
we will generate new data and feed them into the model.

\subsection{Imagery Data}
\label{section:image_data}

The second part of the data
are the images of the locations,
and this part of the data contains all the known wildfire location images
 that was captured by Google Earth Pro\cite{googleearthpro}.
Google Earth Pro provides satellite images for each location,
however, these images are not captured continuously.
For instance, the time when
some images are taken in the same location may have a several-month time difference.
Moreover, the obtainable images may not be the exact date when a fire occurred.
Within the data set, the image adopted will be the first available image
before the wildfire discovery date ($disc\_date\_pre$).
%%XX

\begin{figure}[!htb]
    \centering
    \includegraphics[height=3.5in]{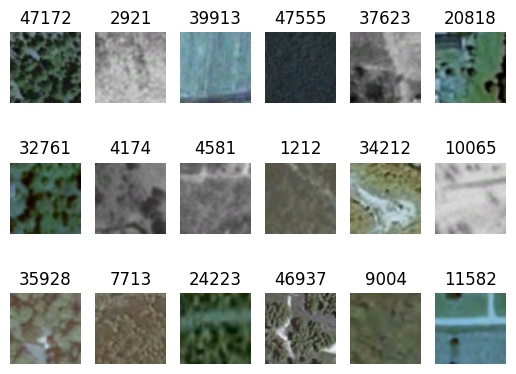}
    \caption{Images randomly chosen from the image dataset,
    and the indices of the images is on top, each image
    is in $100\ pixels \times 100\ pixels$ (Height $\times$ Width)}
    \label{fig:sample_images}
\end{figure}

Moreover, the image qualities in different pictures
are also different due to the different cameras on different satellites
used taking those pictures,
or due to other causes such as different sunlight or cloud conditions.
These images were taken by the cameras or sensors from multiple satellites such as Landsat,
NASA satellite \cite{nasa}, or Sentinel satellites from ESA\cite{esa}
at different times with different sensor technologies.
For example, the most recent United States Geological Survey (USGS)
Landsat 9 has ultra
blue (coastal aerosol) surface reflectance for Band 1,
which allows for coastal watercolor observation,
and Band 2, 3, and 4 surface reflectance are blue, green, and red.
However, some images were taken by using USGS Landsat 7
that only have blue, green, and red bands.
Therefore, the quality may vary due to the improvement of satellite and
sensor technologies.
Furthermore, since the data were taken in 24 years,
most images obtained in the 1990s are black-white grayscale images.
This can be seen from image number 5536 in Figure \ref{fig:sample_images},
and some images may be also in poor quality
due to the glare or sunshine,
such as the number 11191 image
taken from the satellite as shown in Figure \ref{fig:sample_images}.
But after 2005, with the advance in satellite image technologies,
the image quality is much better than those images taken before.
Table \ref{tab:gearth_setting} shows some basic settings when
 the images are captured from Google Earth Pro.
Turning the setting sun off will result in a mask alleviating
the effects of the sun, which can also help keep the consistency of the image quality.
In Google Earth Engine\cite{googleearthpro}, it automatically masks
the clouds if the cloud is in the satellite image, thus, one can obtain
ground images without clouds by unchecking the cloud in the settings.
The map scale is in default for Google Earth Pro, when the images
are being processed and it will crop and re-scale the size of the images.
The road or any other irrelevant marks or pins are also removed when capturing the images.
Thus, as shown in Figure \ref{fig:sample_images}, the images taken
from Google Earth Pro are purely terrain-type without any pins or marks.

%%XX
%%
\begin{table}[!htb]
    \centering
    \begin{tabular}{|c|c|c|c|}
        \hline
        Parameters & Value\\
        \hline
        \hline
        Scale &  1:1000\\
        \hline
        Historic imagery & on\\
        \hline
        Terrain & on\\
        \hline
        Cloud & off\\
        \hline
        Atmosphere & off \\
        \hline
        Sun & off \\
        \hline
    \end{tabular}
    \caption{Settings for Google Earth Pro}
    \label{tab:gearth_setting}
\end{table}

To efficiently generate data, automated software is used to generate the image data.
With the help of Pulover's Marco Creator\cite{pulover}, a software that
can automate simple mouse clicks and complex macros with loops and conditions\cite{pulover}
 is used, which can automatically
capture the images from Google Earth Pro.
This software can record the action that one makes, then repeat the process until
a certain amount of images are obtained.
In this work, we installed the software and Google Earth Pro in a Dell Precision
Tower 5820 desktop,
and we collected all image data in almost three months.
After we gathered the images, we cropped and scaled the images into the size
or scale that desired, which is H: $100\ pixels \times \text{W: } 100\ pixels$
within the approximate length of 100 meters by 100 meters where the center
is the coordinate of the wildfire occurred.

%%XXXX

%%

To obtain 1.88 million images from Google Earth is time-consuming even
using automation software. Each image entry may take more than half
a minute to finish, not including time delay caused by
other reasons such as data requesting and data fetching.
Sometimes, some locations do not have satellite images, which
will lead the computer to pause because of the incorrect
date format due to the software's default settings.

\subsection{Data Pre-processing}

To maintain the consistency of image data information
and avoid poor contrast, when handling the image data,
the data were transformed and normalized \cite{normalize}
all the images to gray-scaled to keep consistency
with the low contrast images, and it will be normalized the image data before training
them by dividing the gray-scaled image data by 255.
After the process, these two data sets are concatenated by the accordance index and generate
a new data set
including the weather data, vegetation, and processed satellite images from the map.
Therefore, the ready-to-train data set has a dimension of 29550
rows and 21 columns plus the image features.

Before training, the data were split into three subsets,
the first one is temperature and wind,
then the second subset includes two more features: humidity and precipitation.
In the third subset, vegetation categories are included to evaluate
the performance of the baseline models with different features.
Our current target is to predict the probability of whether
a certain natural condition will cause wildfire at certain locations,
for instance lightning, as a main ignition source of natural wildfire
\cite{perez2023}.
Even though some wildfire machine learning models in the literature
also claim to be able to predict human-caused fire occurrence, e.g.
\cite{CostafredaAumedes2017HumancausedFO},
predicting human-caused wildfire is not the objective of the present work.

\begin{figure}[!htb]
    \centering
    \begin{minipage}{0.5\textwidth}
        \centering
        \includegraphics[width=2.5in]{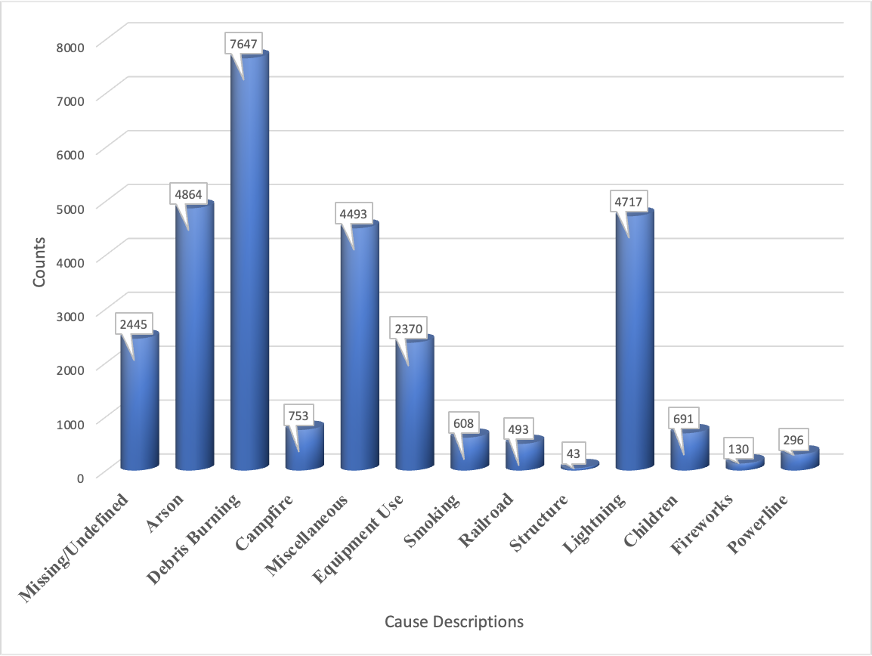}
        \begin{center}
        (a)
        \end{center}
    \end{minipage}\hfill
    \begin{minipage}{0.5\textwidth}
        \centering
        \includegraphics[width=2.3in]{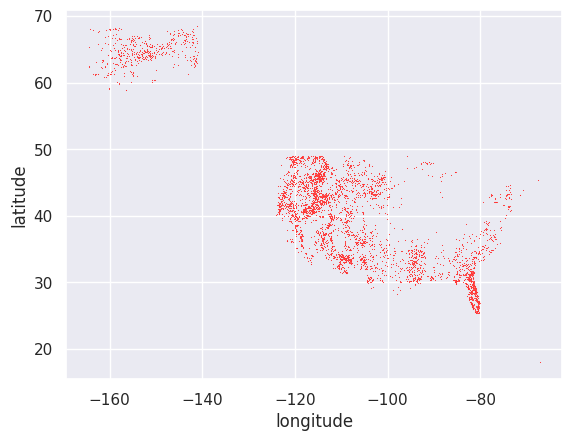}
         \begin{center}
        (b)
        \end{center}
    \end{minipage}
    \caption{
    (a) Histogram of different wildfire causes, and (b)
    Natural caused (lighting) wildfire distributions.}
\label{fig:HistogramX}
\end{figure}

Other causes shown in \ref{nwcg_causes}
are also directly or partially directly caused by humanities,
for now, it is hard to predict them
because they are based on human behaviors \cite{Sean2022},
whose statistical information is not clear.
While the cause of power line may not be caused by human activities\cite{Yao_2022},
thus in the data description one can see that the relative frequency of wildfire
caused by power lines has an extremely small amount
compared to that of other causes, as shown from Figure \ref{fig:HistogramX} (a).
For this perspective,
as shown in Figure \ref{tab:nwcg}, one can treat
this problem as a classification problem.
Thus, for predicting wildfires caused by nature,
lightning can be used as a label.
However, the proportion of the label
as lightning has about $4717$ cases among $29550$,
as shown in Figure \ref{fig:HistogramX} (a), which is only about $15.96\%$
of the entire data.
From Figure \ref{fig:HistogramX} (b) that the data points are much more
sparse compared to the entire data point in Figure \ref{fig:firemap} (b),
which may lead the data imbalanced \cite{4633969}.
Since about $84\%$ of the binary label is false,
and it will have a significant effect on the prediction results,
which the model will tend to have more false in results instead of true,
which can lead to a low true positive rate,
because identifying all false will lead to an average $84\%$ accuracy.
Thus, a re-sampling method could resolve this issue,
which will be discussed in the subsection follows.

\subsubsection{Data Resampling}
There are several data re-sampling methods, e.g. undersampling\cite{4633969},
oversampling or SMOTE\cite{Chawla_2002}.
In the proposed model, two methods are applied: One is Synthetic Minority,
an oversampling technique (SMOTE)\cite{Chawla_2002},  that increases minority by analyzing
the minority population and generate new samples to increase the amount instead
of simply copying the sample to data, which may cause an overfitting problem (see \ref{apd:smote}).

\begin{figure}[!htb]
    \centering
    \includegraphics{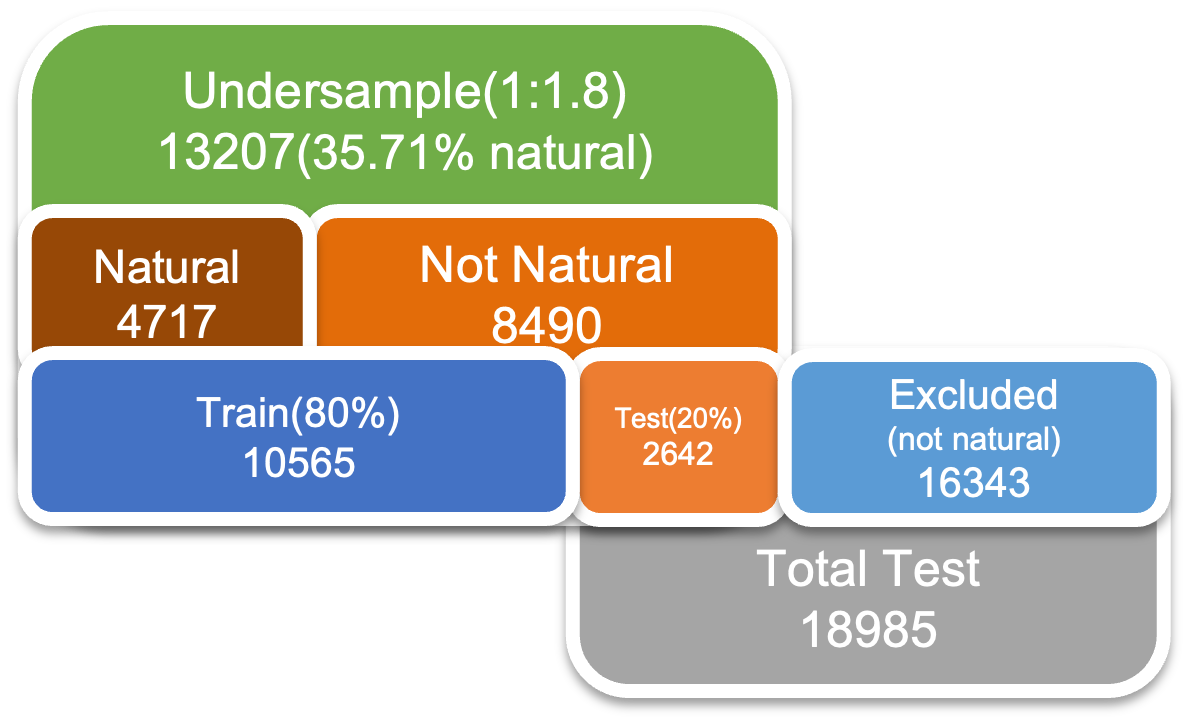}
    \caption{Data split detail, Total 29550, naturally caused 4717, non-natural
    caused 8490, train size 10565, test size 2642, total test 18985}
    \label{fig:train_test_dat}
\end{figure}

The other method used is the undersampling\cite{4633969} method.
This method simply decreases the proportions of the majorities
by deleting occurrences of the more frequent class.
It can balance uneven datasets by keeping all of the data in the minority
class and decreasing the size of the majority class\cite{4633969}.
Nevertheless, it may decrease the train data size.
In the present work, this method is used by randomly choosing
a certain amount of data from majorities and combining them with the minority to adjust the proportion.
By putting the extra majorities to the test case, these methods increase
the size of the test cases.
However, it helps us check the performance of the model,
since most of the data are not being trained, and test data are even
larger than train data.
Therefore, between these two methods, we favor the undersampling approach.
Thus, re-sampling the amount of the minority data to a certain level,
say, the odds can be $1:1.8$, which is about $35.72\%$
as shown in Figure \ref{fig:train_test_dat} is an optimal choice.

Therefore, we randomly choose data from non-natural causes,
then combine it with $4717$
nature-caused wildfire data to keep the proportion as $35.72\%$.
If we set the proportion to $1:1$,
which is $50\%$ for the training and the test data,
it will lead to a very small training
data set with $7547$ data trained and a larger test set.
Thus, we prefer an optimal solution of setting the proportion.
Through this method, the test set will be relatively large
compared even to the train set,
it can help us test the model performance better.

Nevertheless, this may decrease the training data size.
In the present work,
we apply the method by randomly choosing
a certain amount of data from majorities and combine
them with the minority to adjust the proportion.
We will put the extra majorities to the test case and increase the size of test case.
This will lead a large amount of test cases, however,
it actually helps us checking the performance of the model,
since most of the data are not being trained, and test data are even large than train data.
Therefore, between these two methods, we favor the undersampling approach.
Thus, we re-sample the amount of the minority data to a certain level,
say, the odds can be $1:1.8$, which is about $35.72\%$ as shown
in Figure \ref{fig:train_test_dat}.

Before training, the data will be first split into two sets: the training set and the test set.
If the proportion of the total data is about $80\%$ of the training set and $20\%$ test set,
it will have a total of $10565$ in the training set.
Based on this setting, the rest of the non-natural causes data will all
be in the test set which leads to $2642+16343=18985$ test cases.
For the training purpose alone,
a small size training data set may not be good, however,
a larger test size can provide more plausible results.
This method also helps to check whether the model is overfitted during training.
Most of the non-natural causes are not included in the model
during training using undersampling.
When testing the model, these cases are the model never seen during training.
Thus, if the model still performs well during testing,
the model is not affected by the proportion, and it will not be overfitted.

\subsection{Data Uncertainties}

\label{limitations}

In data processing and characterization,
we assume that all the data provided
by the source is accurate enough including the images
from Google Earth Engine. It contains less epistemic uncertainty,
if any. In reality, however, this may not be the case.

In general, outliers may not affect the machine learning training results,
however, some data features that have significantly large
error may affect
the training results.
For instance, in the first part of the data, the weather condition
information at a specific location
may be recorded from the nearest National Weather Service Station (NWS)
or the nearest National Oceanic and Atmospheric Administration (NOAA) station, while
some stations may be far away from the location,
Thus, some of the weather information may not be accurate.
Moreover, the date and time when the fire was discovered may also
not be accurate.
Wildfire is a complex process and its occurrence is the product
of several interrelated factors, including ignition source, fuel composition, weather,
and terrain topography feature \cite{doi:10.1139/er-2020-0019}.
When a fire starts, especially for those nature-caused fires,
it may take a while for people to find or discover them,
if the fire is far away from society.
Under those circumstances, the causes of the fires
may also be hard to find out,
therefore, some causes may also not be accurate.

As mentioned in Section \ref{section:image_data},
the qualities of the images are varied due to the technologies.
Another assumption is also made that the area before
a wildfire occurred may not change a lot during that period.
The images are stitched from Google Earth Engine, and all pieces of
an image are not consequently captured by satellites at the same time.
Some images taken even have several years difference, for example,
 the last image for a certain location may be captured two years ago.
Furthermore, the season when the images are being taken
may also affect the results.
Based on the observation of the images, satellite images
do not provide normal qualities of some locations such as some place
in Alaska or some places that hardly have people, which means
that the image has quite low resolution,
and some images only appeared with less color spectrum
without showing the exact terrain.
Therefore, the quality of image data still needs to improve.

Moreover, the data used in this work only contain
the data of the wildfire that had occurred inside the United States.
Thus, based these data,
the prediction of the wildfire occurrence outside the U.S. may not be accurate.

\subsection{Data and Code Availability}

It is both preferable and sustainable to feed more data to train the AI model to obtain better results.
All training data can be found on the official websites, and imagery data is also available on \ref{colab_instruct}.
The computer codes for generating visualization are available on \href{https://tinyurl.com/2p9ezkzm}{Colab}.
See \ref{colab_instruct} for the ReadMe and instructions.

\section{Methods}

\subsection{Statistical Settings}
The technical objective of this work is to predict whether a location
is at risk of wildfire occurrence. The prediction is characterized
by a set of conditions, natural or man-made. Mathematically,
this prediction may be expressed
in terms of conditional probability defined as follows,
\begin{align}
    \label{eq:target}
    \mathbb{P}(wildfire|natural\ causes\textcolor{black}{,\ other\ conditions})
\end{align}

In our data format,  all the data entries are related to the wildfires
that have been actually
occurred.
By virtue of probability chain rule,
the condition probability of a wildfire occurrent due to natural causes, or
other conditions can be expressed as follows,
\begin{eqnarray}
    \label{eq:prob_data}
    &&\mathbb{P}(natural\ causes|\textcolor{black}{wildfire,\ other\ conditions})
    \nonumber
    \\
    &=&
    \frac{\mathbb{P}(wildfire, natural\ causes\textcolor{black}{,\ other\ conditions})}{\mathbb{P}(wildfire\textcolor{black}{,\ other\ conditions})},
\end{eqnarray}

Finally, we can reformulate Eq. (\ref{eq:target}) by using Bayes' Theorem,
\begin{eqnarray}
   && \mathbb{P}(wildfire|natural\ causes\textcolor{black}{,\ other\ conditions})=
    \nonumber
    \\
    && \frac{\mathbb{P}(natural\ causes|wildfire\textcolor{black}{,\ other\
    conditions})\mathbb{P}(wildfire\textcolor{black}{,\ other\ conditions})}{\mathbb{P}(natural\
    causes\textcolor{black}{,\ other\ conditions})}=
    \nonumber
    \\
    && \frac{\mathbb{P}(natural\ causes|wildfire\textcolor{black}{,\
    other\ conditions})\mathbb{P}(wildfire|\textcolor{black}{other\ conditions})
    \mathbb{P}(\textcolor{black}{other\ conditions})}{\mathbb{P}(natural\
    causes\textcolor{black}{,\ other\ conditions})}
    \nonumber
    \\
    \label{eq:math_model}
\end{eqnarray}

When we are modeling for $\mathbb{P}(natural\ causes|wildfire\textcolor{black}{,\ other\ condiions})$,
we assume that the location has had a wildfire and try to identify the conditional probability of its cause.
Among these probabilities, the probabilities, $\mathbb{P}(natural\ causes\textcolor{black}{,\ other\ conditions})$ \\
and $\mathbb{P}(other\ conditions)$ shown in Eq. (\ref{eq:math_model}),
can be predicted by \textcolor{black}{other sources such as} weather forecasts.
Then based on the prediction of weather forecast, we can obtain
the probabilities due to natural causes, i.e. the probabilities of storm weather in a certain area.
For the $\mathbb{P}(wildfire|\textcolor{black}{other\ conditions})$,
we can refer from Probability of Ignition\cite{prob_ignition} provided by NWCG.
Instead of directly predicting the conditional probability,
our model will focus on the prediction of
the likelihood probability of the natural causes given the occurrence of wildfire,
for instance, the lightening caused wildfire,
which can be directly found in the data.
Thus, once we know three of them, we can have all the probability needed to
make a posterior prediction.

Furthermore, based on Eq. (\ref{eq:math_model}), we have
\begin{align*}
& \mathbb{P}(wildfire|natural\ causes,\ other\ conditions)
\\
& \propto
\mathbb{P}(natural\ causes|wildfire,\ other\ conditions),
\end{align*}
Thus if one of the probabilities increases
the other one will act the same,
and we can choose one to represent the other.

\begin{figure}[!htb]
    \centering
    \includegraphics[width=6in]{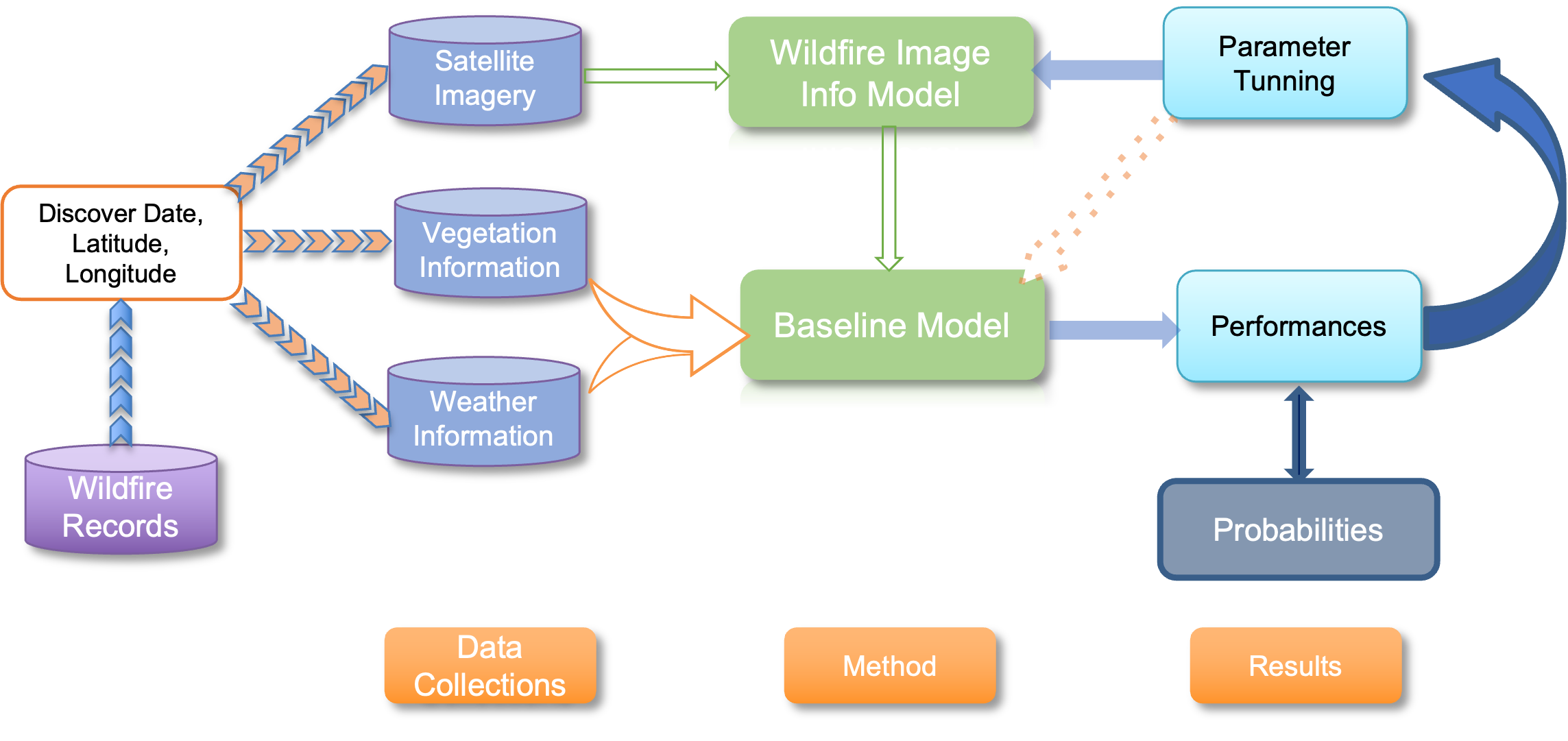}
    \caption{Overview of Data Flow in Model}
    \label{fig:method_outline}
\end{figure}

Figure \ref{fig:method_outline} is a flowchart that outlines data flow for the proposed model discussed in the previous Section.
For each past wildfire record, we collected the weather, location, date, vegetation,
and image data based on the given location and date.
For the information data without images, which is the features associated with locations,
weather, and vegetation categories, we process the data directly to a neural network\cite{6953413}.
Then our model without images can also provide predicted probabilities as outputs.
To validate the correctness of the proposed model, we defined a classification threshold, $0.5$.
Thus, if the probability is less than the threshold, we categorize it as false otherwise 1.
With these metrics, we can assess the accuracy of the proposed model.
For the data with images, inspired by architectures of CNN\cite{oshea2015introduction}, ResNet\cite{he2015deep}, and Transformer\cite{vaswani2017attention},
we designed and implemented a feature extraction model that consists of
a light version of ResNet with a modified ViT\cite{dosovitskiy2021image} with image features
to exaggerate the features that it can discern but humans cannot
see by naked eyes and with as much low computation cost as possible.
By doing so, we can then output some new features to predict the risks
of wildfire at certain locations.

Moreover, to have a stable results,
the model will first find the mean of two probabilities and then
predict the probabilities in Eq. (\ref{eq:prob_data}).
To quantify the probabilities,
the confusion matrix is used to determine whether the results
predicted by the model are accurate or not.

As shown from Figure \ref{fig:method_outline},
all the features in information data and image data are based
on the location and weather conditions.
Hence, with some feature selections, a designed model is used
to extract and enlarge the image features, so that
we can predict the probabilities such as
\begin{align*}
    \mathbb{P}(wildfire|natural\ causes\textcolor{blue}{,\ other\ conditions})
\end{align*}
For the given location and weather information,
the transformer machine learning model can obtain the data
from the sources and calculate the probabilities.
If the probabilities are larger than the threshold that we set,
we can then say that if a given natural cause occurs
the fire should occur under the circumstance of the weather
and vegetation/terrain conditions.
Otherwise, the fire will not occur.

\subsection{Baseline Model}

In the first phase of the training,
the baseline model used is a deep learning neural network\cite{LeCun15},
in which only the information data is used to illustrate how new features would affect the results.
We used several different combinations of features to observe the result or output variations,
which include temperature plus wind, temperature plus wind plus humidity plus precipitation,
temperature plus wind plus humidity plus precipitation plus hot-encoded vegetation, etc.
First, only geo-referenced records would be fed into the model.
Both the classification method and regression method are used to predict the probability results.
Since in this case, one would like to have a probability result
and the regression strategy can provide better and more stable results,
thus, regression strategy is used for the task,
and the results shall be discussed in the next section.

\begin{figure}[!htb]
    \centering
    \includegraphics[width=\textwidth]{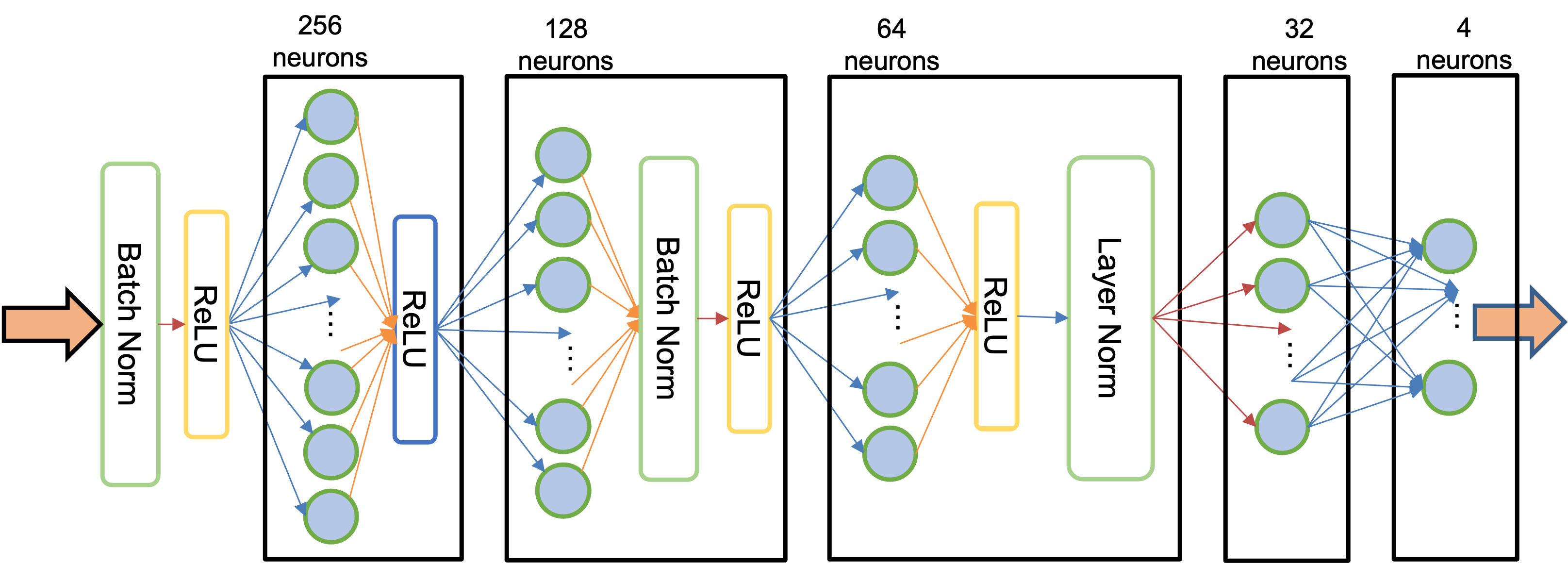}
    \caption{The architecture of the proposed machine learning model
    as a hybrid deep neural network.}
    \label{fig:simplenn_architecture}
\end{figure}

For this neural network architecture,
the model has five fully connected hidden layers,
and each layer can be treated as a block.
As shown in Figure \ref{fig:simplenn_architecture},
after the input layer, each of these consists of two or three
parts, hidden layer, batch normalization layer \cite{pmlr-v37-ioffe15}.
Here, the activation function is rectified linear unit (ReLU)\cite{agarap2019deep},
while the last layer is another fully connected layer.
Inside this model, the total amount of variables is $49705$.

From the batch normalization algorithm, through
a linear combination, the mean and standard deviation of each input
for the block will be within a certain range,
so that the next layer does not need to satisfy the changes and increase
the efficiency of the model\cite{santurkar2019does}.
Moreover, the model will depend less on hyper-parameter tuning
so that it allows a little larger learning rate.
It will also shrink the internal covariance shift\cite{pmlr-v37-ioffe15}.

From the batch normalization algorithm, we can see that through
a linear combinations, the mean and standard deviation of each input
for the block will be within a certain range,
so that the next layer does not need to be adjusted with
possible changes and increase
the efficiency of the model\cite{santurkar2019does}.
Moreover, the model will depend less on hyper-parameter tuning
so that it allows a little larger learning rate.
It will also shrink the internal covariance shift\cite{pmlr-v37-ioffe15}.

Since we do not use batches in this model, the batch norm layer will be
simply the normalization of the entire input matrices.
With the same architecture shown in Figure \ref{fig:simplenn_architecture},
we appended vegetation categories and trained the model again.

Inside the model, as shown in Table \ref{tab:nn_params},
the loss function we use is the mean square error (MSE),
which is a wildly used loss function, and
the Cross Entropy loss\cite{NEURIPS2018_f2925f97}
is also available in the code for classification.
\begin{align}
    MSE = \frac{\sum^{n}_{i=1} (y_{i} - \hat{y}_{i})^{2}}{n}
\end{align}
where $n$ is total amount of data, $y_{i}$ is the $i$th true label, and $\hat{y}_{i}$
is the $i$th predicted label by model.
This metric provides an intuitive difference between the true values
and the predicted values.

Another way is treating the model as a classification problem.
Thus in the model, the loss function can be the Cross-Entropy loss function \cite{10.5555/1146355},
 which requires one-hot encoding of the label and changing
 the layer in the baseline model to SoftMax\cite{GoodBengCour16}.
A new loss function called true rate loss will calculate
both sensitivity and specificity are defined, and then maximized it.
For finding the gradient, the Adam optimizer\cite{kingma2017adam} is used.

\begin{table}[!htb]
    \centering
    \begin{tabular}{c|c}
        \hline
        Parameters & Values\\
        \hline
        Hidden Layers & [256, 128, 64, 32, 4]\\
        \hline
        Loss Function & Mean Square Error (Cross Entropy) \\
        \hline
        Optimizer & Adam optimizer\cite{kingma2017adam}\\
        \hline
        Learning Rate &  0.01\\
        \hline
        Weight Decay & 1e-4\\
        \hline
        Epochs & 100 \\
        \hline
        Batch Size & NA \\
        \hline
    \end{tabular}
    \caption{Hyperparameters for Neural Network}
    \label{tab:nn_params}
\end{table}

Training this model does not necessarily need a local GPU and large RAMs.
It would only take about several seconds for 100 epochs without a mini-batch
by using a standard NVIDIA T4 Tensor Core GPU with 16G RAM
on Google Colabotory\cite{colab}, where the codes
and pre-trained models are also available
on given \href{https://tinyurl.com/2p9ezkzm}{Colab}.
For a single CPU training without a mini-batch method,
it will also only take less than $30$ seconds.

\subsection{Wildfire Imagery Information Net (WIIN)}

Besides the information data, for the imagery data collected,
we designed a Wildfire Image Information Net, which can extract
the features from the images according to our purpose.
It is a combination of a deep learning architecture of
ResNet\cite{he2015deep} type
and a transformer with an attention-based architecture.

Initially, a transformer is used in translation for
the natural language process,
thus, the self-attention mechanism focuses more on the locations
of a word and its score in a given sentence,
which is beneficial for understanding complex language structures.

When applying the transformer to images, a
conventional vision transformer
focuses on classification or segmentation.
It represents the information by token features,
which are also known as class features.
However, our model is not concerned
with classification but only interested in the information
embedded in images,
Therefore, the proposed model focuses more
on what has been ``told'' by an image.

\begin{figure}[!htb]
    \centering
    \includegraphics[width=5.5in]{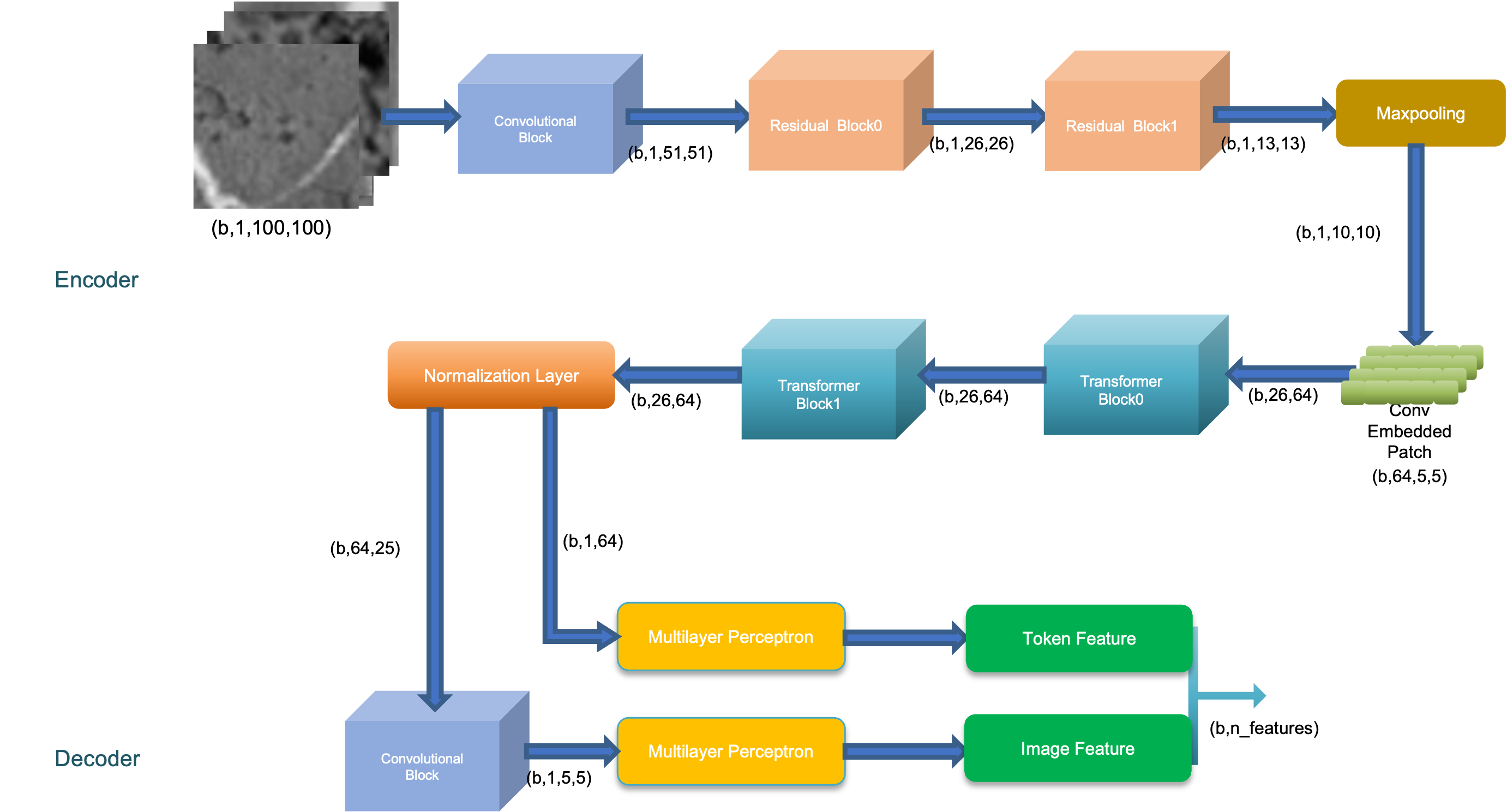}\\
    \caption{Architecture of Wildfire Imageries Information Net (WIIN) (b:batch size)}
    \label{fig:encoder}
\end{figure}

To make this work for the abandoned image pixels from the
conventional vision transformer,
the model not only applies the self-attention mechanism but also
adds a convolutional block to process the image pixels through
a multilayer perceptron so that the model can extract
image features from it as shown in Figure \ref{fig:encoder}.

The reason why the ResNet is applied before the transformer\cite{vaswani2017attention}
is because usually, a transformer requires a large amount of data\cite{9944625}.
However, in our case, instead of using a large amount of data
like GPTs\cite{radford2018improving}
are using,
we have limited training data size, after the undersampling method
is applied, we only have about $10000$
data for training.
Thus, in order for the transformer can have better performance in image processing,
we enlarge the features by using ResNet.
ResNet has CNN-like\cite{oshea2015introduction} architecture,
it can extract the feature maps from the original image\cite{cnnfeature},
which would make it easier for a transformer to process
the feature maps instead of the original images.
Moreover, based on its architecture, a transformer with attention
mechanisms will act as an encoder to analyze these images.
After embedding an image, it can assign the scores and
find the most possible patches that are related to the target.
As shown in Figure \ref{fig:encoder}, we split the output features into two parts:
the token feature and the image feature.
Both features will append to the information data
as new features and feed into
a deep neural network for training. Thus, with the original 21 features,
there are a total of 23 features when using the combined model.

\subsubsection{Residual Network}

\begin{figure}[!htb]
    \centering
    \begin{minipage}{0.45\textwidth}
        \centering
        \includegraphics[width=\columnwidth]{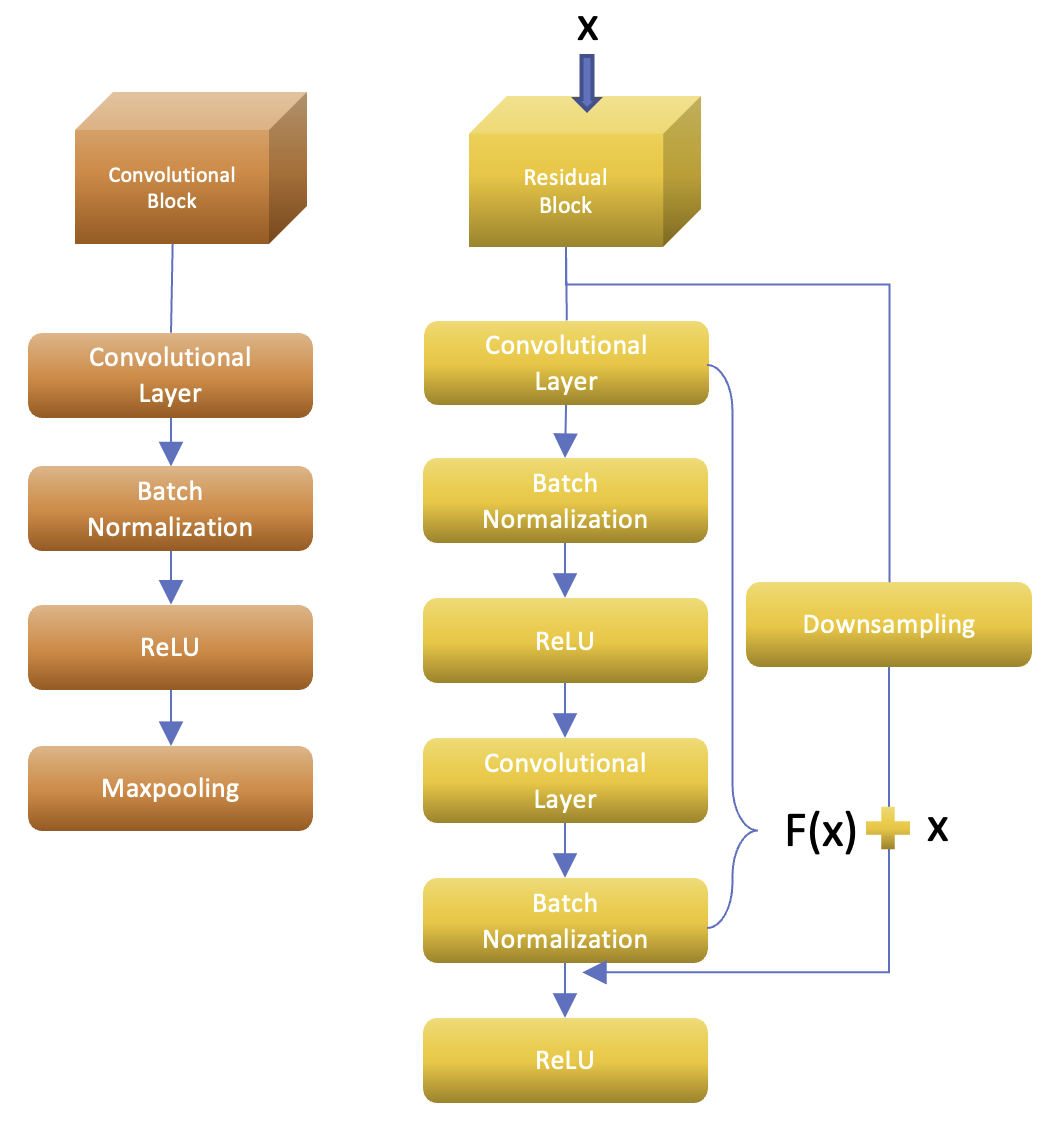}
       \begin{center}
       (a)
       \end{center}
    \end{minipage}\hfill
    \begin{minipage}{0.45\textwidth}
        \centering
        \includegraphics[width=\columnwidth]{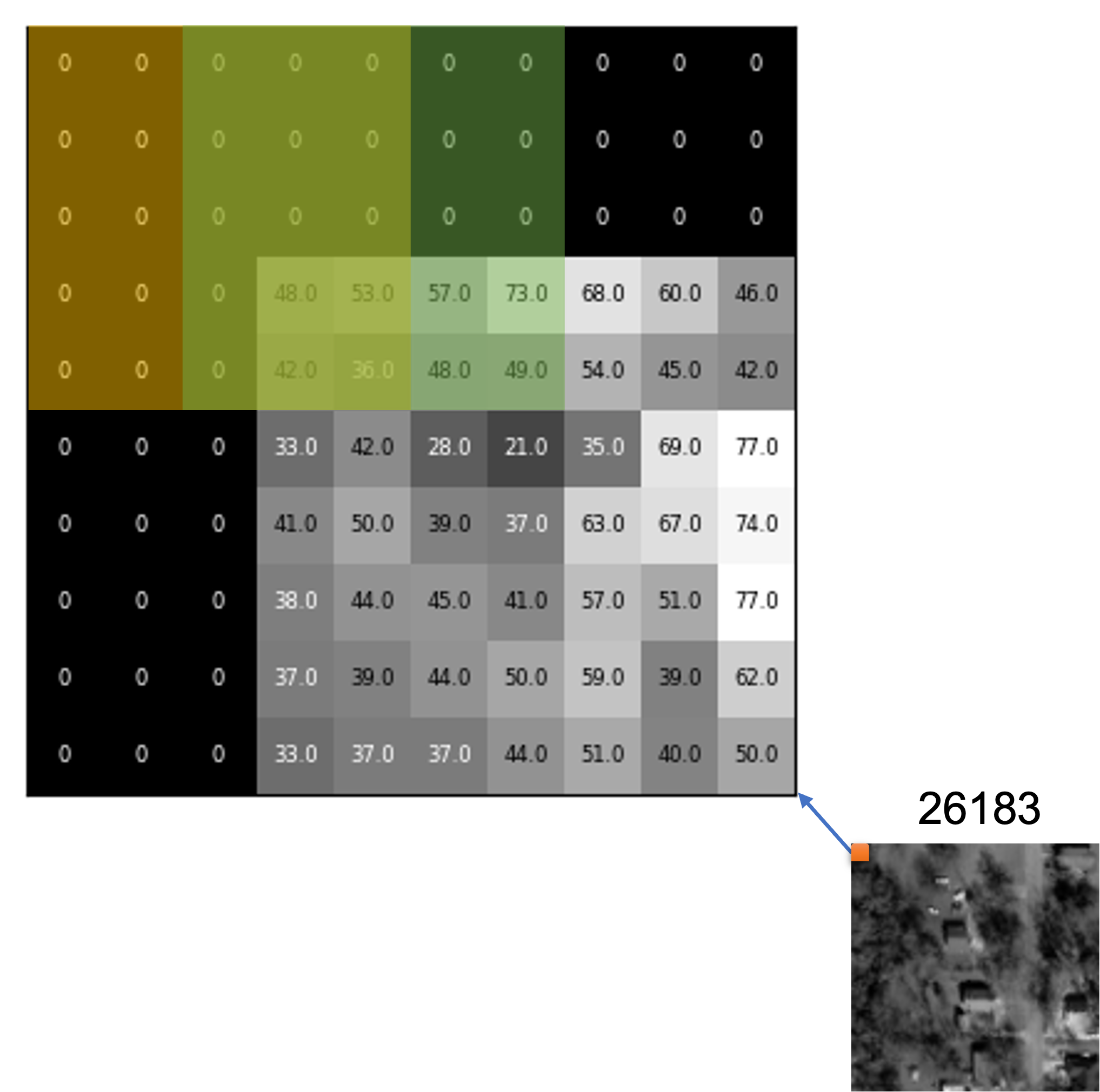}
        \begin{center}
       (b)
       \end{center}
    \end{minipage}
    \caption{(a) Convolutional Block; RIGHT: Residual Block, and (b)
    Top left corner of image 26183 in the first convolutional
        layer with kernel $ 5 \times 5 $, stride $2$, padding $3$.
    }
    \label{fig:Inner}
\end{figure}

At this stage, feature maps from the original captured image are extracted
so that the model can both maintain its performance and save computation power.
As shown from Figure \ref{fig:encoder}, the mini-batched images
with the size of $ (100, 100) $ will first go through a convolutional block,
which includes a convolutional layer, a batch normalization layer, an activation layer,
and a max-pooling layer, as shown on the left of Figure \ref{fig:Inner} (a).
The convolutional layers are used to extract features
from any previous procedures, and it will output a new feature map of the previous
stage of an image or feature map \cite{cnnfeature}.
Mathematically, for a grey-scaled image $I \in \mathbb{R}^{W \times H \times 1}$,
where W stands for the width of the image, and H is the height of the image,
the ij-th entry of output feature map by using the kernel $K \in \mathbb{R}^{p \times q}$
can be represented as  \cite{berzal2019redes}
\[
    \sum_{i}\sum_{j}K_{(i,j)}I_{(p-i,q-j)}
\]

In this stage, the convolutional layer has a kernel size of 5, stride of 2, and padding of 3.
Taking Figure \ref{fig:Inner} (b) as an example,
the blackboxes with 0s are the padding area, and the rest are pixels of the original image.
The $ 5 \times 5 $ yellow and green shaded areas will
multiply weighted kernels that will be trained at different times,
and each time, the kernel will move two boxes since the stride is set to 2.

Similar to the convolutional layer, the maxpooling layer will keep the main features
and reduce parameters in the model, it removes all the values except
for the maximum value in each filter process \cite{cnnfeature}.
It will then obtain the maximum numbers within each kernel instead
of multiplying the trained weights.
Therefore, after the convolutional block, the image size
will become $ (51, 51) $ as unnecessary features would be removed.

Then, the convoluted images will be fed into two residual blocks with three and two layers separately.
The residual blocks are mainly two convolutional blocks without the max-pooling layer,
added with downsampled initial inputs as residuals if necessary.
As shown in right of Figure \ref{fig:Inner} (a),
for a single residual block, the new residual $ F(x) $ will add back to
initial input value $ x $ to create a new output $ F(x) + x $.
With the help of adding residuals to the output, ResNet
can provide a good feature map from the original image without
losing too much information and prevent the gradient vanish or exploding \cite {ren2016object,he2015deep}.
Even if the fed data has a small image size, it works well.
After two residual blocks, the image size reduces to $ (13, 13) $,
and we then feed the processed images into the last max-pooling layer.
Mathematically, let $F(\cdot) = BN(Conv(\cdot))$
Similar to the convolutional layer, the max-pooling layer will keep the main features
and reduce parameters in the model, it removes all the values except
for the maximum value in each filter process \cite{cnnfeature}.
It will then obtain the maximum numbers within each kernel instead
of multiplying the trained weights.
Therefore, after the convolutional block, the image size
will become $ (51, 51) $ as unnecessary features would be removed.

Then, the convoluted images will be fed into two residual blocks with three and two layers separately.
The residual blocks are mainly two convolutional blocks without the max-pooling layer,
added with downsampled initial inputs as residuals if necessary.
As shown in right of Figure \ref{fig:Inner} (a),
for a single residual block, the new residual $ F(x) $ will add back
to initial input value $ x $ to create a new output $ F(x) + x $.
With the help of adding residuals to the output, ResNet
can provide a good feature map from the original image without
losing too much information and prevent the gradient vanish or exploding \cite {ren2016object,he2015deep}.
Even if the fed data has a small image size, it works well.
After two residual blocks, the image size reduces to $ (13, 13) $,
and we then feed the processed images into the last max-pooling layer.
Mathematically, let $F(\cdot) = BN(Conv(\cdot))$
\begin{align*}
    z_{0} &= F(x)\\
    z_{0} &= maxpool(g(z_{0}))\\
    z'_{l} &= F(g(z_{l-1}))\\
    z_{l} &= g(F(z'_{l}) + z_{l-1})\\
    z_{l} &= z_{l} + \sum^{L-1}_{i=1}F(z_{i})\\
    y &= maxpool(z_{l})
\end{align*}
where $BN$ stands for the batch normalization layer, $Conv$ is the convolutional layer,
$maxpool$ is the maxpooling layer, and $g(\cdot)$ is the activation function.
\begin{table}[!htb]
    \centering
    \begin{tabular}{p{0.4\linewidth} | p{0.285\linewidth}}
        \hline
        Parameters & Values\\
        \hline
        Layer(Kernel, Stride, Padding) & conv(5,2,3); maxpool(3,2,1); conv(3,1,1)+conv(3,1,1)$\times$3; conv(3,2,1)+conv(3,2,1)$\times$2; maxpool(4,1,0)\\
        \hline
        % Loss Function & Mean Square Error \\
        % \hline
        % Optimizer & Adam optimizer\cite{kingma2017adam}\\
        % \hline
        % Learning Rate &  0.01\\
        % \hline
        % Weight Decay & 1e-4\\
        % \hline
        Epochs & 300 \\
        \hline
        Batch Size & 32 \\
        \hline
    \end{tabular}
    \caption{Hyperparameters for ResNet}
    \label{tab:resnet_params}
\end{table}

\subsubsection{Wildfire Images Transformer (WIT)}
\begin{figure}[!htb]
    \centering
    \includegraphics[width=.8\columnwidth]{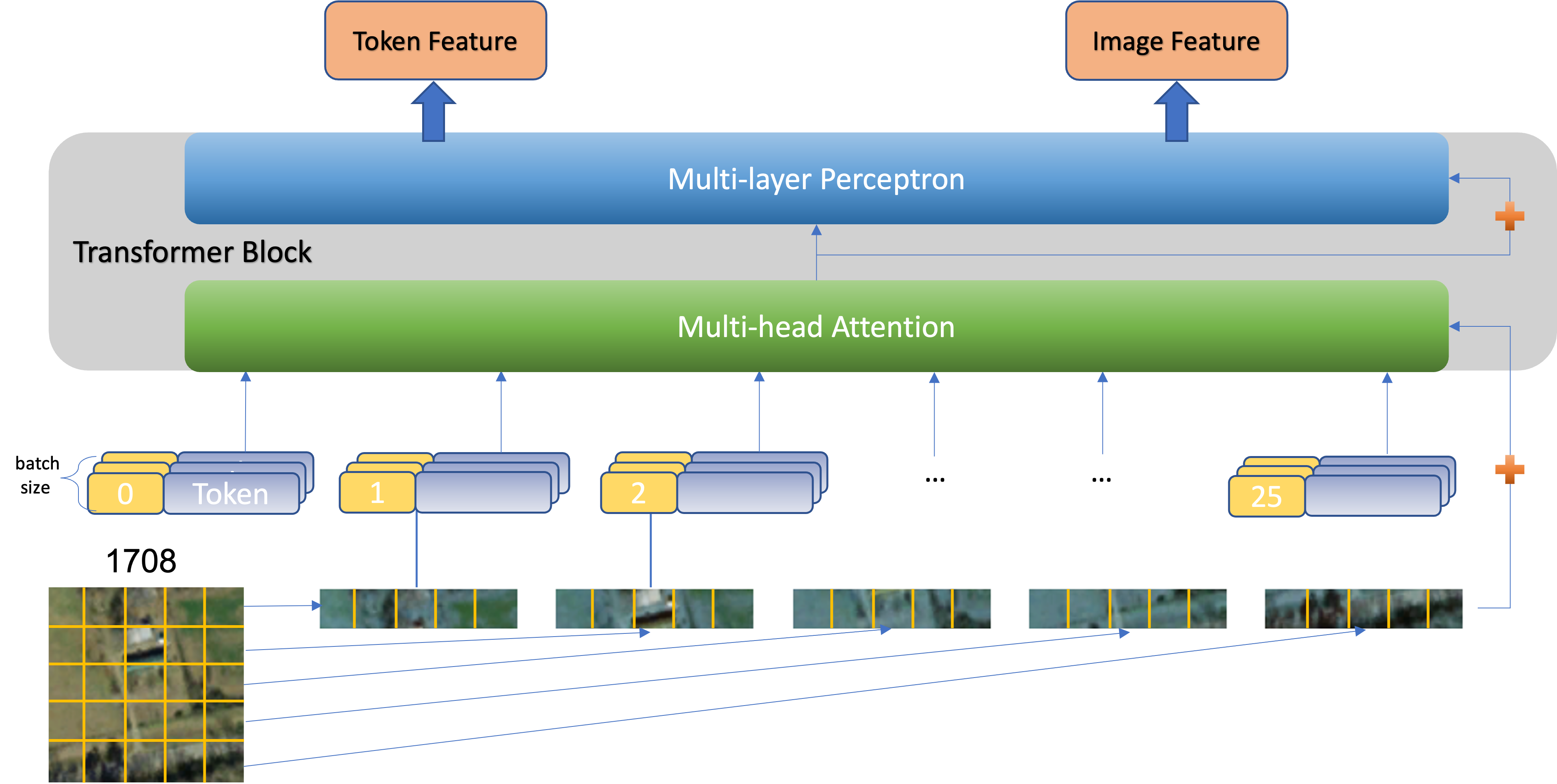}
    \caption{Vision Transformer}
    \label{fig:vit}
\end{figure}

\begin{figure}[!htb]
    \centering
    \includegraphics[width=4.5in]{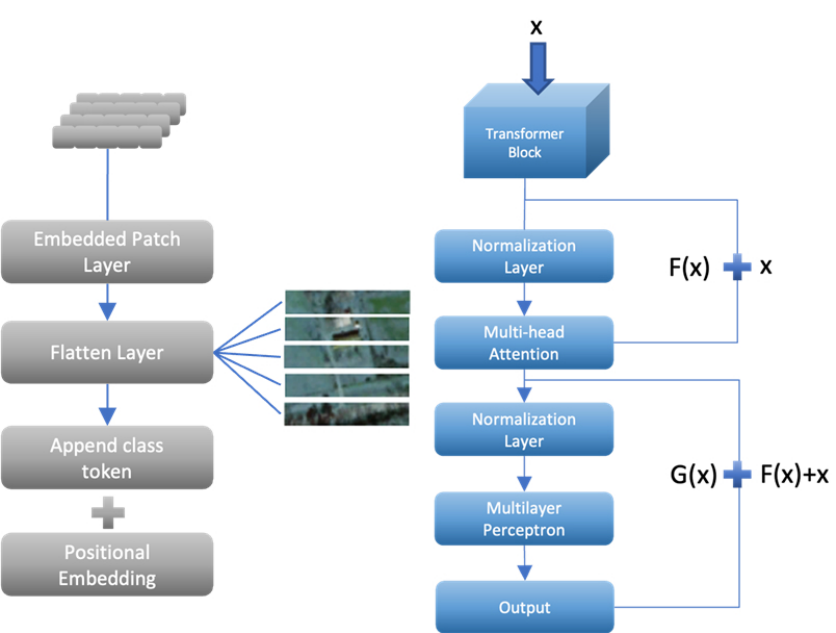}
    \caption{LEFT: Embedded Patch Layer; RIGHT: Transformer Block}
    \label{fig:inner_transformer}
\end{figure}

In the transformer stage, similar to a vision transformer (ViT) (See: \cite{dosovitskiy2021image}),
the output image from the previous stage will be fed into a convoluted embedded patch layer so that
the image features will be flattened, as shown in Figure \ref{fig:vit}, meanwhile,
a learnable feature token will be appended to the newly flattened features, as shown in the left of
Figure \ref{fig:inner_transformer}.
Mathematically, this block with the output shown in Figure \ref{fig:vit} can represented as
\begin{align*}
    z_{0} &= [x^{0}_{token};x^{1}_{patch}E;x^{2}_{patch}E; \cdots ;x^{N}_{patch}E]+E_{position}\\
    z'_{l} &= MSA(LN(z_{l-1}))+z_{l-1}\\
    z_{l} &= MLP(LN(z'_{l}))+z'_{l}\\
    y_{image} &= Linear(Conv(z^{1:N}_{l}))\\
    y_{token} & = Linear(LN(z^{0}_{l}))\\
\end{align*}
where $x^{i}_{patch}$ is the patches, $ N $ is number of patches,
which can be calculated as $\frac{H \times W}{(patch\ size(P))^{2}}$ $ E $s
are the linear projection of flattened patches, a fully connected layer,
which will map the learned features onto sample space (See: \cite{cnnfeature}),
$E_{pos}$ is the positional encoder,
$MSA$ is the multi-head self-attention layer, $LN$ is the linear normalization layer,
$MLP$ is the multi-layer perception layer,
$linear$ is the fully connected linear layer, and $ conv $ is the convolutional layer.
For instance, for an image in the output of ResNet, $ x_{i} \in \mathbb{R}^{1 \times 10 \times 10 }$,
with the patch size of 2, the flattened vector should have $
\frac{H \times W}{P^{2}} = \frac{100}{4} = 25$ patches concatenate an additional
token vector and positional encoding.
The output dimension will become a $ x \in \mathbb{R}^{(25+1) \times 4}$ features.
Our model applied a convolution method to the image since our patch size is small,
only $4$, with the output channel $ 64 $.
Thus, the output dimension of the embedded patch layer is $ (26, 64) $.

Then, the data will have two transformer blocks, each of which will
have a multi-head attention layer and a multilayer perception, as shown
in the right of Figure \ref{fig:inner_transformer}.
In the multi-head attention layer, the flattened input feature vectors
 with appending tokens will apply linear projections by multiplying three
different weight, $w_{i} \in
\mathbb{R}^{P^{2}C \times \frac{P^{2}C}{n\_head}}, i \in \{q,k,v\}$
to three new vectors $Q(query)$, $K(key)$, and $V(value)$, for instance $w_{q} x = Q$.
Thus, in this case, Query, key, and value vectors should have
shape $ (b, 26, 8, 8)$ as the hidden dimension is $64(8\times8)$.
The $n\_head$ stands for the number of heads defined for multi-head attention.
Vaswani et al.\cite{vaswani2017attention}. define the output of
the attention layer as concatenating the various heads
and then multiplying by a matrix $w^{o}$.

\begin{align*}
    \text{concat}(head_1, \cdots, head_h)w^{o} = head_1 w^{o}_{1} + \cdots + head_h w^{o}_{h}
\end{align*}
where $w^{o} = \begin{bmatrix} w^{o}_{1}\\ \vdots\\ w^{o}_{h}\end{bmatrix}$
As the results are concatenated, the dimension is $ \frac{P^{2}C}{n\_head} \times n\_head = P^{2}C $.

Thus, at the end of this layer, the dimension is still
$ \mathbb{R}^{P^{2}C \times P^{2}C} $.
Once $ Q, K, V $ are obtained, the score could be computed by $softmax(QK^{T})$,
and logit are scaled by $\frac{1}{\sqrt{d_{qkv}}}$ for
the scaled dot-product score as follows,
\[
    Z = softmax(\frac{Q K^{T}}{\sqrt{d_{qkv}}})V,
\]
where $Z$ is scaled dot-product attention\cite{vaswani2017attention}.

During multi-layer perception stage, the input will flow into
a full connected linear layer with activation function Gaussian
Error Linear Units(GELUs)\cite{hendrycks2020gaussian} which is defined as
\begin{align*}
    \text{GELU}(x) &= xP(X \leq x)\\
    &= x \cdot \frac{1}{2}[1+\text{erf}(\frac{x}{\sqrt{x}})]
\end{align*}
and a dropout layer to prevent overfitting.
In the last, linear layer the output dimension will be changed back to its origin.
Thus the dimension of the vectors has no changes.

Instead of outputting a token class only like ViT, our model will also output image features,
which can be treated as summarized features of an image.
Thus, at the output stage of the wildfire image information network,
we will have extra features for appending to the neural network model.
\begin{table}[!htb]
    \centering
    \begin{tabular}{c|c}
        \hline
        Parameters & Values\\
        \hline
        Number of Layers & 2\\
        \hline
        Patch Size & 2 \\
        \hline
        Number of Head & 8\\
        \hline
        Multi-Layer Perceptron Neurons & 8 \\ %32\\
        \hline
        Dropout Rate & 0.2\\
        \hline
    \end{tabular}
    \caption{Hyperparameters for Transformer}
    \label{tab:transformer_params}
\end{table}

\subsection{Hybrid Multimodal Model}

\begin{figure}[!htb]
    \centering
    \includegraphics[width=6in]{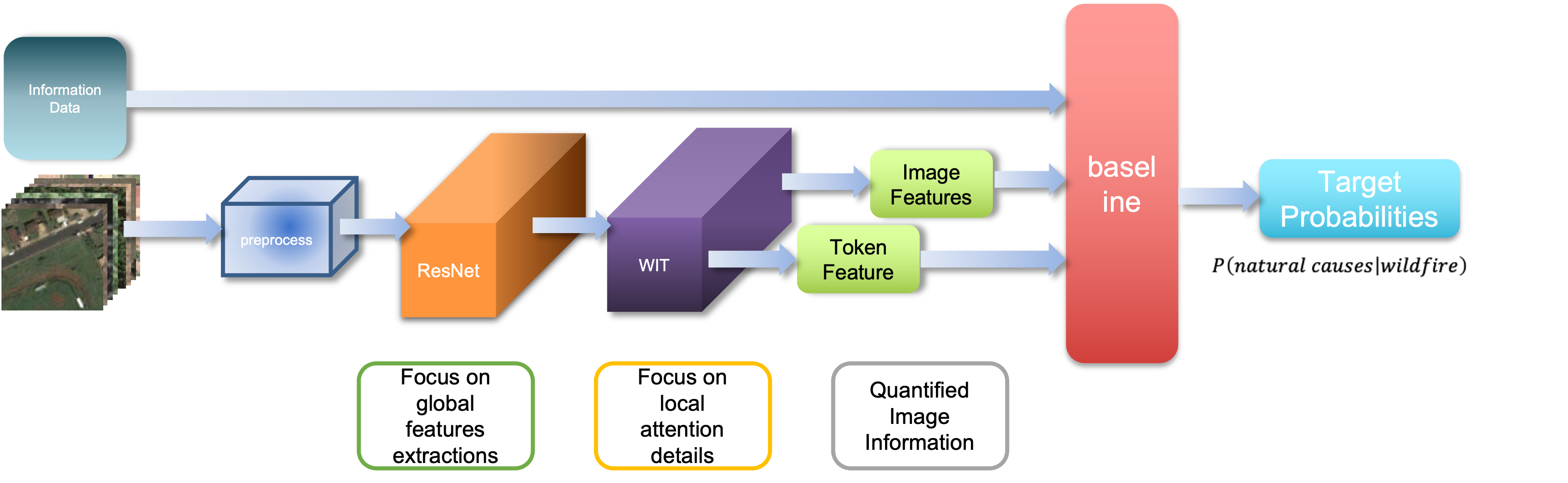}
    \caption{Hybrid Model Outline}
    \label{fig:method_outline2}
\end{figure}

In the hybrid model, the imagery data will be processed by the previous model
and image features and token features will be the outputs from the process.
Then the outputs will concatenate with the information data and fed the new concatenated
data into the baseline model.
Thus, as shown from Figure \ref{fig:method_outline2}, the hybrid model
will take the results from wildfire images information net, then together
with the information data into the baseline model.
At the last stage, the output of the model will be the required target probabilities.
In the metrics, the threshold is $0.5$, which means if the probability
is larger than $0.5$, then the model is categorized as true or false.
Mathematically, we can write
\begin{align*}
    \hat{R} =
    \begin{cases}
        True, & \text{if}\ p > 0.5\\
        False, & \text{if}\ p \leq 0.5
    \end{cases}
\end{align*}
where $\hat{R}$ are the responses, $p$ is the outputs from the model.
During the training, it usually reaches the top $TPR$ and $TNR$ at about $100$ epochs.
In the hybrid model, the model totally has $268615$ variables.

\section{Results and Discussions}

\subsection{Forecast Metrics}

The metric used to measure the performance is the True Positive Rate
(TPR or sensitivity) and true negative rate (TNR or selectivity) \cite{FAWCETT2006861},
which both are the components of accuracy.
True positive is defined as values that are positive and predicted positive.
In this case, the True Positive value indicates a scenario in which a wildfire that occurred in a given location was caused by natural causes, and the prediction also indicates that the wildfire is caused by natural causes. Thus, this value should be as large as possible.
On the contrary, a False Negative value represents a scenario in which a wildfire occurred in a given location, and
the model predicts that it is due to natural causes, while they are not. This is a type II error; thus, it should be avoided.
Therefore, a true positive rate with these two values can provide us with good performance metrics, Since the model tends to have a high true negative rate without using the undersampling method, finding optimal values between the true positive rate and the true negative rate is a top priority.
It provides a conditional probability that if there is a wildfire,
the model can categorize whether it is caused by nature or not.

The metric used to measure the performance is the True Positive Rate
(TPR or sensitivity) and True Negative Rate (TNR or selectivity) \cite{FAWCETT2006861},
which both are the components of accuracy.
True positive is defined as values that are positive and predicted positive.
In this case, the True Positive value indicates a scenario in which natural
causes caused a wildfire in a given location, and the model's prediction
also indicates that natural causes caused the wildfire.
Thus, this value should be as significant as possible.
On the contrary, a False Negative value represents a scenario in which
a wildfire occurred in a given location, and
the model predicts that it is due to natural causes, while they are not.
This is a type II error; thus, it should be avoided.
Therefore, a True Positive Rate with these two values can give us
good performance metrics. Since the model tends to have a high true negative
rate without the undersampling method, finding optimal
values between the True Positive Rate and the True Negative Rate is a top priority.
It provides a conditional probability that if there is a wildfire,
the model can categorize whether it is caused by nature.
Mathematically, we have
\begin{align}
    \label{eq:tpr}
    TPR &= \frac{True\ Positives}{True\ Positive + False\ Negatives}\\
    \label{eq:fpr}
    TNR &= \frac{True\ Negatives}{True\ Negatives + False\ Positives}
\end{align}

There are other metrics often used in the analysis, F-score \cite{10.1007/978-3-540-31865-1_25},
which can be represented as
\begin{align*}
    {\rm F{-score}} &= \frac{True\ Positives}{True\ Positives+ 0.5(False\ Positive + False\ Negative)}
\end{align*}
and the precision or the positive predictive value, i.e.
\begin{align*}
    Precision = \frac{TP}{TP+FP}~.
\end{align*}
However, F-score and precision are not included here because, in our test case,
the false counts are much more than the true ones, as shown in the equation.
The F-score gives equal importance to precision and recall \cite{hand1084928040}.

The precision is related to the proportion of true
cases and false cases \cite{10.1145/1143844.1143874},
(See Appendix \ref{apd:precision}).
In our case, since precision will lead to some waste of resources,
our top priorities are still $TPR$ and $TNR$.

Using the terminology coined in this paper,
as can be seen from Eq. (\ref{eq:tpr}),
a high ``True Positive'' rate means that
when the model predicts a fire in a certain area, there is actually
a fire occurs, while the false negative indicates a low rate of
``True Positive'', which means the possibility of
misclassifying the real occurred wildfire.

The term ``False Positive''
in our terminology, means that
there is no fire, and the model correctly categorizes it
as no fire occurrence.
If this rate is low, it indicates that there will be some false alerts
as shown in Eq. (\ref{eq:fpr}),
which may waste resources while ignoring the alert might be detrimental.
Thus, currently, our focus will be first to increase the $TPR$, the $TNR$ and
hence the accuracy in reliability.

\subsection{Baseline Model Results}
\begin{figure}[!htb]
    \centering
    \begin{minipage}{0.49\textwidth}
        \centering
        \includegraphics[width=\columnwidth]{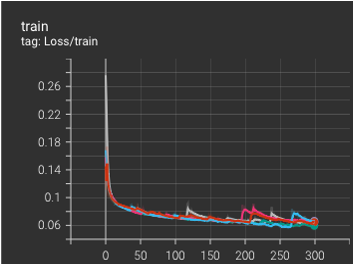}
        \begin{center}
        (a)
        \end{center}
    \end{minipage}\hfill
    \begin{minipage}{0.49\textwidth}
        \centering
        \includegraphics[width=\columnwidth]{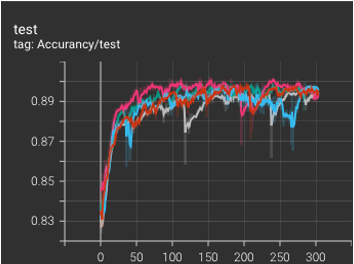}
         \begin{center}
        (b)
        \end{center}
    \end{minipage}
    \caption{ (a) Loss for data without vegetation category, and (b)
    Accuracy for data without vegetation category.
    }
    \label{fig:Convege+LA}
\end{figure}

\begin{figure}[!htb]
    \centering
    \begin{minipage}{0.49\textwidth}
        \centering
        \includegraphics[width=\columnwidth]{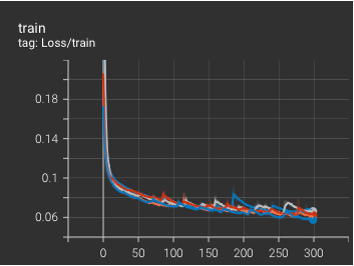}
        \begin{center}
        (a)
        \end{center}
    \end{minipage}\hfill
    \begin{minipage}{0.49\textwidth}
        \centering
        \includegraphics[width=\columnwidth]{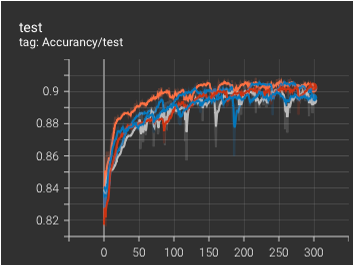}
        \begin{center}
        (b)
        \end{center}
    \end{minipage}
    \caption{(a) Loss for data with vegetation category, and
    (b) Accuracy for data with vegetation category.
    }
    \label{fig:Vege_LA}
\end{figure}

\begin{figure}[!htb]
    \centering
    \begin{minipage}{0.49\textwidth}
        \centering
        \includegraphics[width=\columnwidth]{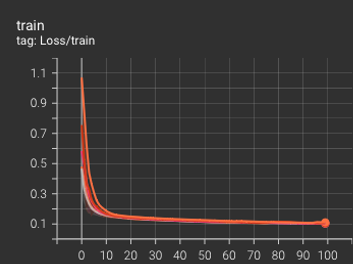}
          \begin{center}
        (b)
        \end{center}
    \end{minipage}\hfill
    \begin{minipage}{0.49\textwidth}
        \centering
        \includegraphics[width=\columnwidth]{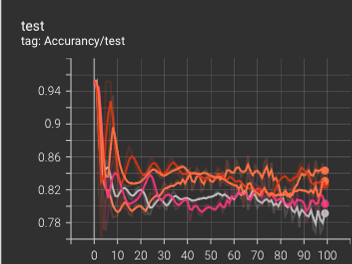}
          \begin{center}
        (b)
        \end{center}
    \end{minipage}
    \caption{
    (a) Loss for undersampling data with vegetation category, and (b)
    Accuracy for undersampling data with vegetation category
    }
    \label{fig:Loss}
\end{figure}

The baseline model with recommended setting (See: \ref{apd:recomm_tune})
provides relatively stable results for all the baseline cases we trained.
As we can see from Figure \ref{fig:Convege+LA} (a), Figure \ref{fig:Vege_LA} (a), and
Figure \ref{fig:Loss} (a), 5 pre-trained results are chosen,
which are also available in \href{https://www.tinyurl.com/2p9ezkzm}{Colab}.
The variance of the losses of each case is relatively low for 5 random runs.
However, with different selected features, the results are slightly improved.

As one can see from Figure \ref{fig:Convege+LA} (a), the losses are successfully converged
to a small but little higher value than $0.06$.
On the accuracy side, as shown in Figure \ref{fig:Convege+LA} (b),
the test accuracy at the last epoch 300 is all within the range of
$89\%$ to $90\%$.

\begin{figure}[!htb]
    \centering
    \begin{minipage}{0.49\textwidth}
        \centering
        \includegraphics[width=\columnwidth]{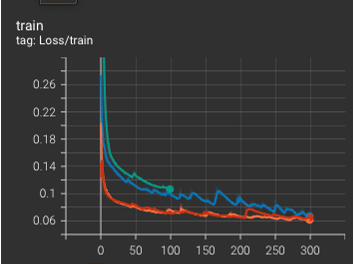}
        \begin{center}
        (a)
        \end{center}
    \end{minipage}\hfill
    \begin{minipage}{0.49\textwidth}
        \centering
        \includegraphics[width=\columnwidth]{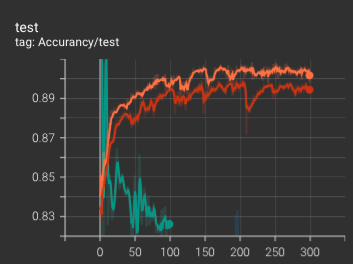}
         \begin{center}
        (b)
        \end{center}
    \end{minipage}
    \caption{
    (a)Comparison of losses, and (b) Comparison of accuracy
    }
    \label{fig:Comparison}
\end{figure}

If we add vegetation categories as new features to the training dataset,
with the same architecture as the hyperparameters, the results will be slightly improved
as shown in Figure \ref{fig:Vege_LA} (a).
Compared with the previous one, the loss is slightly smaller, mostly within 0.06,
as shown in Figure \ref{fig:Comparison}, the line in red is the result of not including
the vegetation category, and the line in orange is the one with the vegetation category.
The same situation occurs in test accuracy as shown in Figure \ref{fig:Vege_LA} (b),
we can see that the overall accuracy is better and some of the results are more than $90\%$ accuracy.
Model Stability is one of the core principles in data science \cite{doi:10.1073/pnas.1901326117}.
For each of the models, we randomly split train and test data, as shown in all errors,
the baseline model is relatively stable.
We can also see from Figure \ref{fig:Comparison} (b) that
the accuracy is even better if the features include the vegetation categories.

As we can see from Figure \ref{fig:Loss} (a) if we use the undersampling data for training,
the losses are higher than previous case, and the accuracies are also lower,
which is around $82\%$.
The reason for the high accuracy is due to the imbalance of data.
The model tends to categorize as false since more than $80\%$ data are false.
Thus, the model fits well for categorizing most cases as false.

\begin{figure}[!htb]
    \centering
    \begin{minipage}{0.495\textwidth}
        \centering
        \includegraphics[width=2.6in]{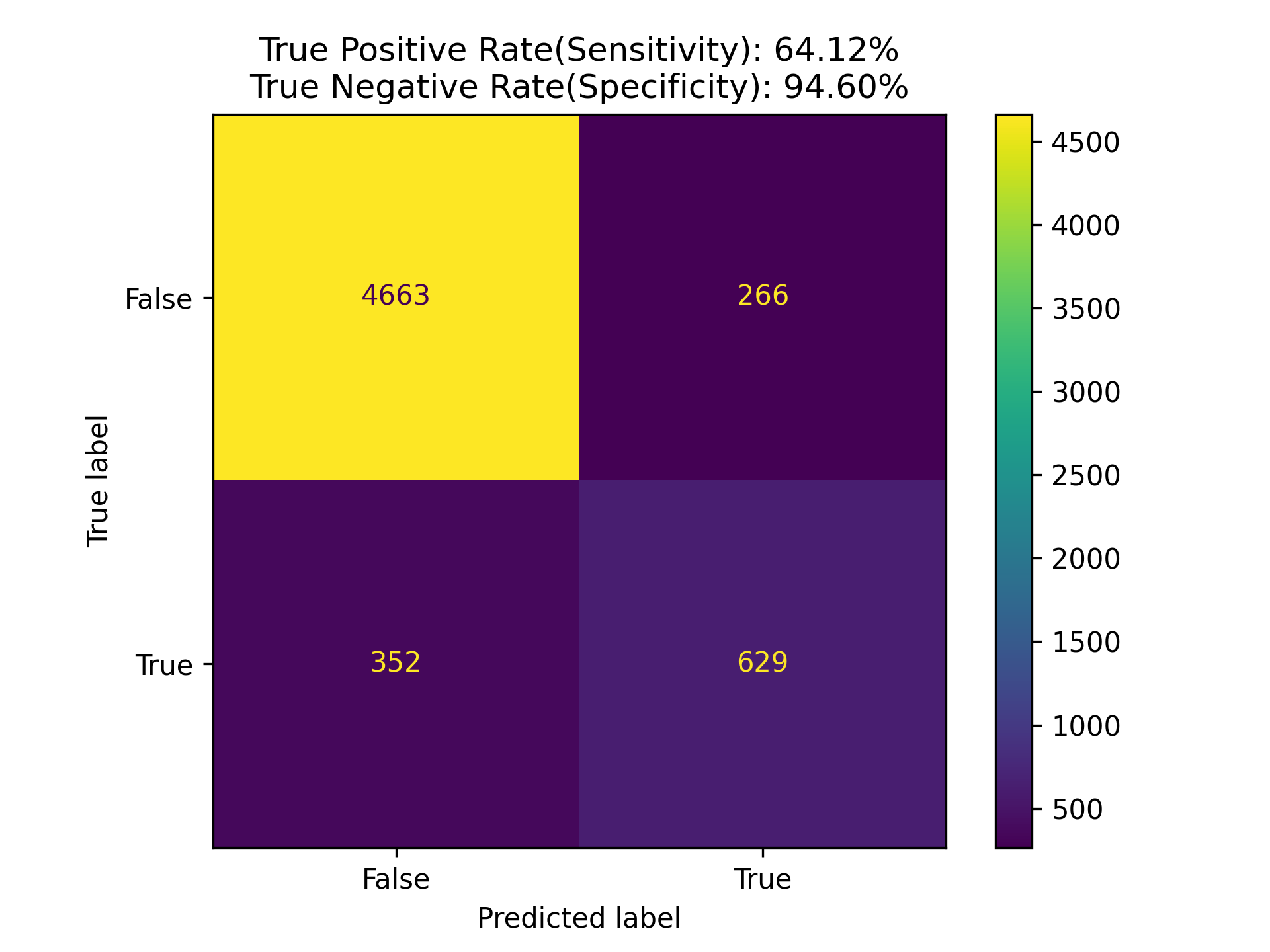}
         \begin{center}
            (a)
            \end{center}
    \end{minipage}\hfill
    \begin{minipage}{0.495\textwidth}
        \centering
        \includegraphics[width=2.6in]{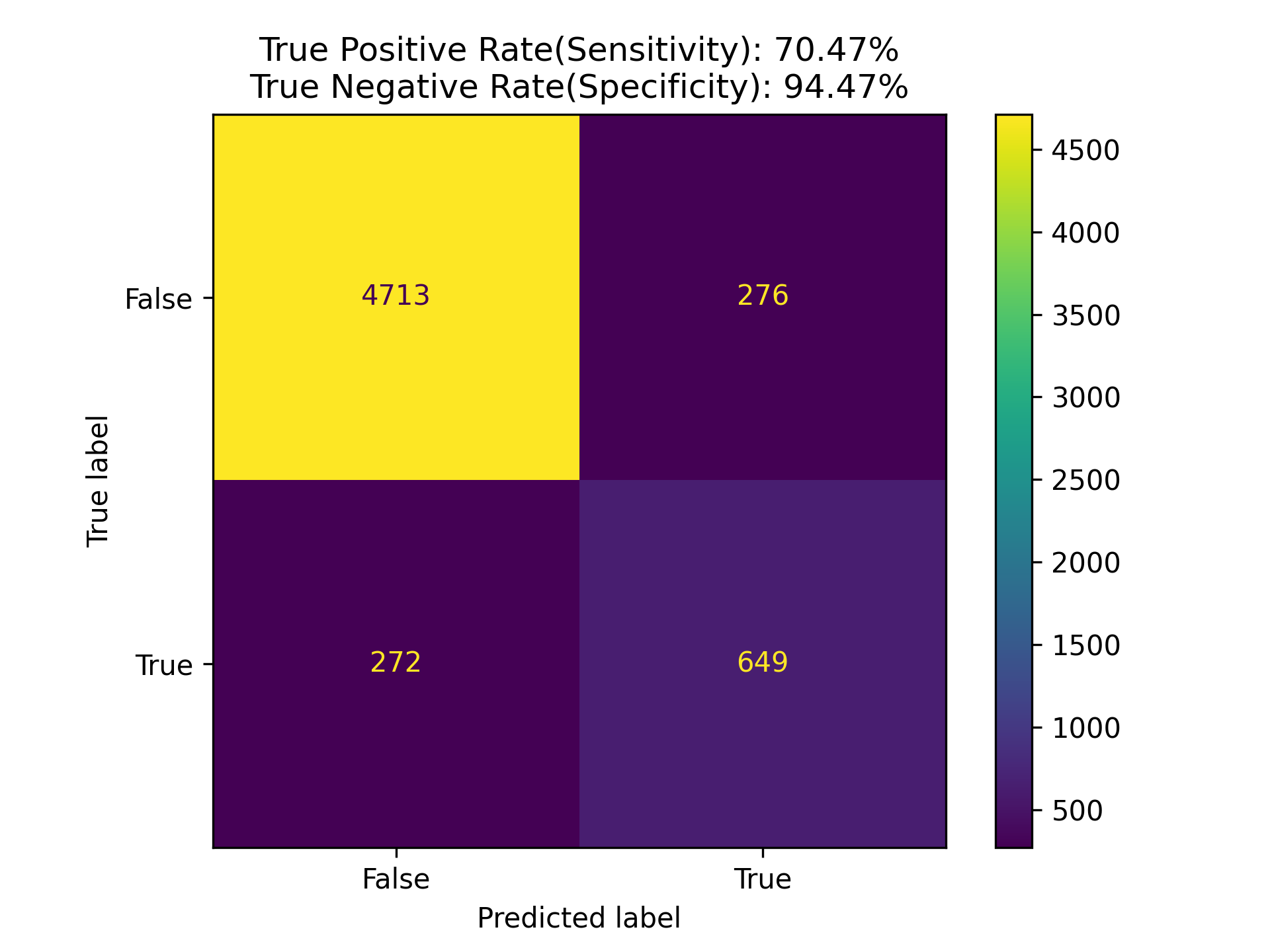}
        \begin{center}
            (b)
            \end{center}
    \end{minipage}\\
     \begin{minipage}{0.8\columnwidth}
        \centering
        \includegraphics[width=2.6in]{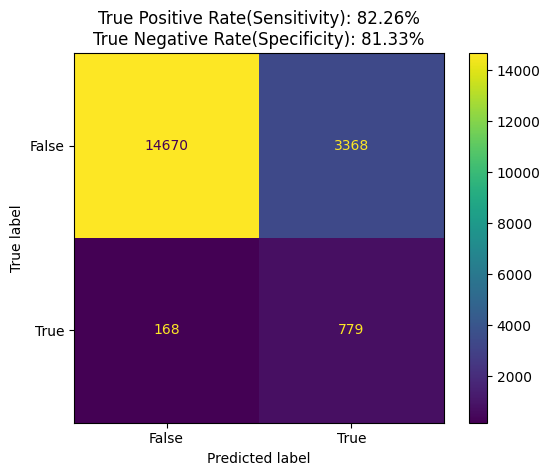}
         \begin{center}
            (c)
            \end{center}
        \end{minipage}
    \caption{
    (a) Confusion matrix for data without vegetation category, (b)
    Confusion matrix for data with vegetation category, and (c)
    Confusion matrix for undersampling model.
    }
     \label{fig:confusion_ABC}
\end{figure}

As shown in Figure \ref{fig:confusion_ABC} (a), we can see that among $5910$ test data,
$5015$ are categorized as negative, which leads to a $94.6\%$ TNR.
Even though it is a little better with vegetation categories,
a similar situation occurs in the cases with vegetation category,
as shown in Figure \ref{fig:confusion_ABC} (b), among the same amount of test data,
$5910$ and $4985$ are categorized as false.
For sensitivity, the results are similar to those without the vegetation category,
which is also more than $90\%$.
For specificity, compared to the previous one, we can see that the results are about $5\%$ better.
Even with limited improvement, these results are better than the previous.
Thus, specificity contributed the most accurately since it has a large proportion and relatively high results.
As we mentioned, sensitivity is the top priority and the part that we care about most,
by using these features, the sensitivity is not high enough even though it provides a high accuracy.
Thus, if we only used 7 days, 15 days, and 30 days' average temperature and wind data as features,
the neural network model converges but with around $89\%$ accuracy and quite low sensitivities ($65\%-70\%$)
with a high specificity with more than $90\%$, as shown in Figure \ref{fig:confusion_ABC} (a).

After applying the undersampling method, we can see the results from
Figure \ref{fig:confusion_ABC} (c), even though the weighted accuracy is lower than in previous cases,
the $TPR$ increases a lot to more than $80\%$.
For a total of $18985$ test cases, the model found an optimal solution
to balance the $TPR$ and $TNR$, which are around $80\%$, leading to the total accuracy also about $80\%$.
This is the desired result.

Then if we add 7 days, 15 days, and 30 days of average humidity or precipitation data,
the neural network model converges, and losses are around 0.07 and accuracy is around $88\%-89\%$,
the confusion matrix is not good since the sensitivities are less than $60\%$.
Once we added vegetation-type data, the losses decreased by about $0.01$ to $0.05-0.06$
and accuracy will increase around $1\%$, the sensitivity of confusion matrix
increases to around more $60\%-65\%$.

\subsection{Hybrid model results}

\begin{figure}[!htb]
    \centering
    \begin{minipage}{0.25\columnwidth}
        \centering
        \text{21850}\\
        \text{88.182\%}\\
        \text{TP}\\
        \includegraphics[width=1in]{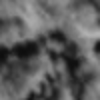}\\
        \includegraphics[width=1in]{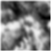}\\
        \includegraphics[width=1in]{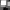}\\
    \end{minipage}\hfill
    \begin{minipage}{0.25\columnwidth}
        \centering
        \text{3107}\\
        \text{17.672\%}\\
        \text{FN}\\
        \includegraphics[width=1in]{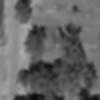}\\
        \includegraphics[width=1in]{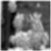}\\
        \includegraphics[width=1in]{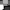}\\
    \end{minipage}\hfill
    \begin{minipage}{0.25\columnwidth}
        \centering
        \text{22587}\\
        \text{0.377\%}\\
        \text{TN}\\
        \includegraphics[width=1in]{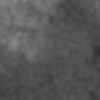}\\
        \includegraphics[width=1in]{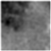}\\
        \includegraphics[width=1in]{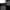}\\
    \end{minipage}\hfill
    \begin{minipage}{0.25\columnwidth}
        \centering
        \text{36704}\\
        \text{62.585\%}\\
        \text{FP}\\
        \includegraphics[width=1in]{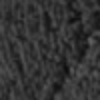}\\
        \includegraphics[width=1in]{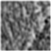}\\
        \includegraphics[width=1in]{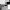}\\
    \end{minipage}
    \caption{Sample outputs of ResNet with their predicted probabilities and confusion matrix categories. TOP: original gray-scaled images ($ H: 100 \times W: 100 $); MIDDLE: outputs of first convolutional layer (first principle component) ($ H: 51 \times W: 51 $); BOTTOM: outputs of last layer ($ H: 10 \times W: 10 $)}
    \label{fig:layer_img}
\end{figure}

Figure \ref{fig:layer_img} shows how some images change during different stages
from a pre-trained model, we can see the top of Figure \ref{fig:layer_img} is the original
grey-scale image
and the output of the first convolutional layer of the same index are the middle images.
After the first trained convolutional layer, the trained results successfully
maintain the features from the original images.
We are able to see the obvious features from the previous original images.
For the bottom of Figure \ref{fig:layer_img}, due to the resolution being quite small,
which is $10 \times 10$, we can see some similarities if compared with the original images.
Thus, when finishing training, the convolutional part extracts
feature maps from the images and feeds them to the next stage.

\begin{figure}[!htb]
    \centering
    \begin{minipage}{0.25\columnwidth}
        \centering
        \text{21850}\\
        \includegraphics[width=1in]{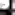}\\
    \end{minipage}\hfill
    \begin{minipage}{0.25\columnwidth}
        \centering
        \text{3107}\\
        \includegraphics[width=1in]{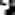}\\
    \end{minipage}\hfill
    \begin{minipage}{0.25\columnwidth}
        \centering
        \text{22587}\\
        \includegraphics[width=1in]{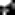}\\
    \end{minipage}\hfill
    \begin{minipage}{0.25\columnwidth}
        \centering
        \text{36704}\\
        \includegraphics[width=1in]{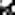}\\
    \end{minipage}
    \caption{Sample outputs of Transformer stage. Images are up-sampled
     by scale factor 3($ H: 15 \times W: 15 $) with bilinear mode}
    \label{fig:translayer_img}
\end{figure}

Figure \ref{fig:translayer_img} is the output of the pre-trained model with the same index from
Figure \ref{fig:layer_img}.
The original images have a resolution of $5 \times 5$, even with a small resolution,
we believe that this is how the model sees
the images and finds the features of a certain condition for the model.
Figure \ref{apd:sample_out} shows detailed outputs for each layer.

\begin{figure}[!htb]
    \centering
    \begin{minipage}{0.5\textwidth}
        \centering
        \includegraphics[width=2.5in]{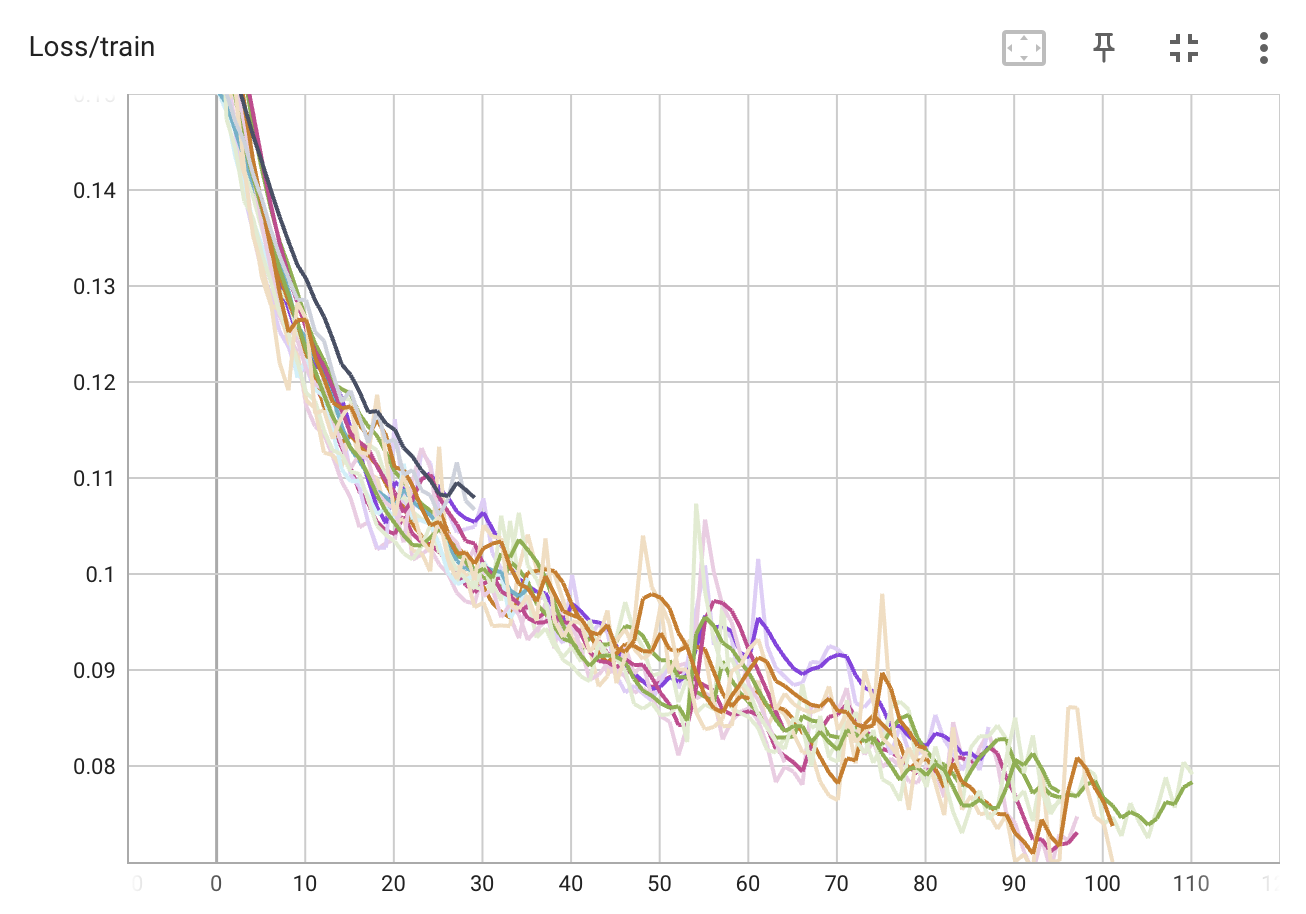}
         \begin{center}
            (a)
            \end{center}
    \end{minipage}\hfill
    \begin{minipage}{0.5\textwidth}
        \centering
        \includegraphics[width=2.5in]{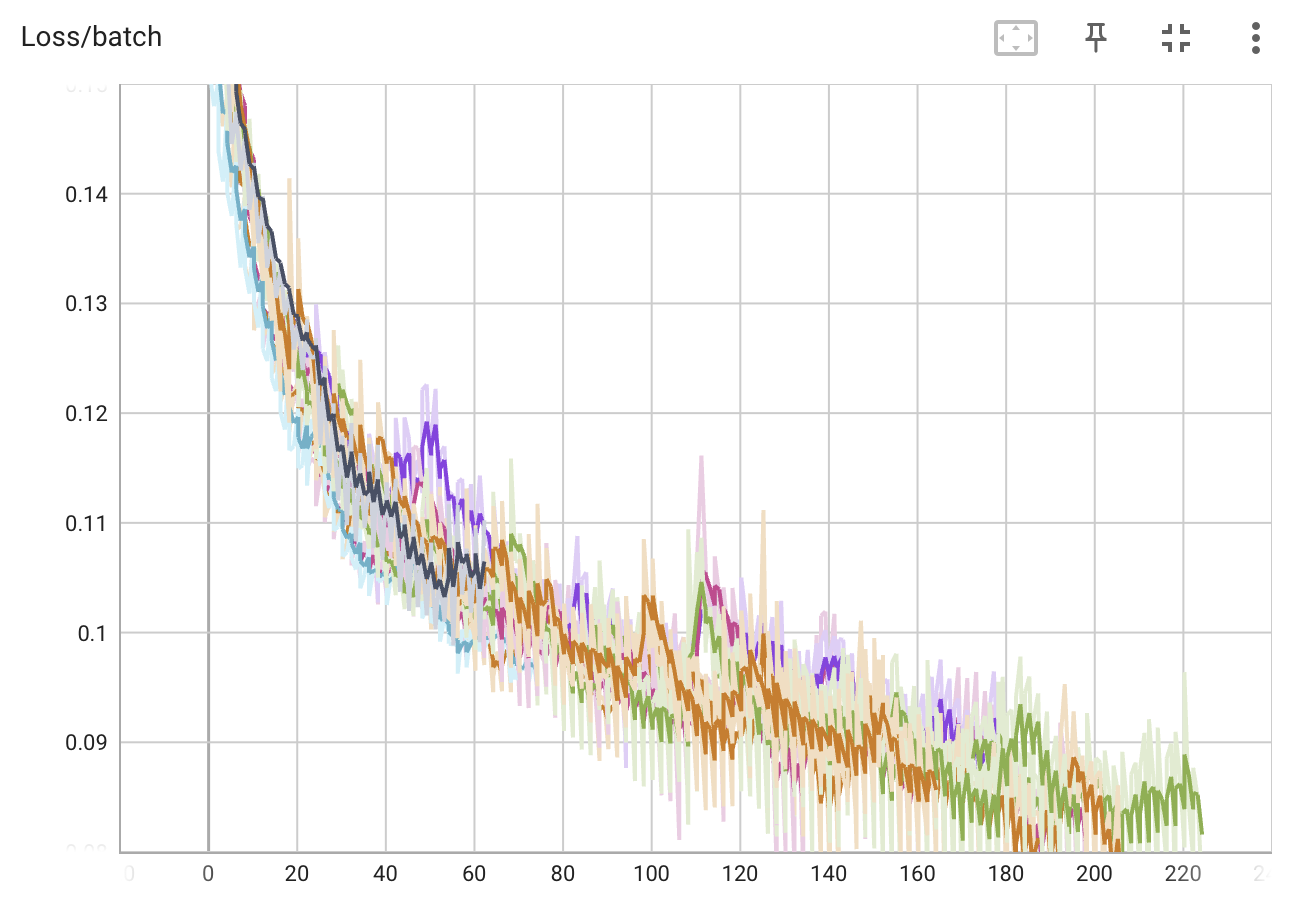}
         \begin{center}
            (b)
            \end{center}
    \end{minipage}
    \caption{(a) Epoch losses (left) and (b) scaled batch losses (right) of hybrid model}
    \label{fig:combined_epoch_loss}
\end{figure}

\begin{figure}[!htb]
    \centering
        \begin{minipage}{0.5\columnwidth}
            \centering
            \includegraphics[width=\columnwidth]{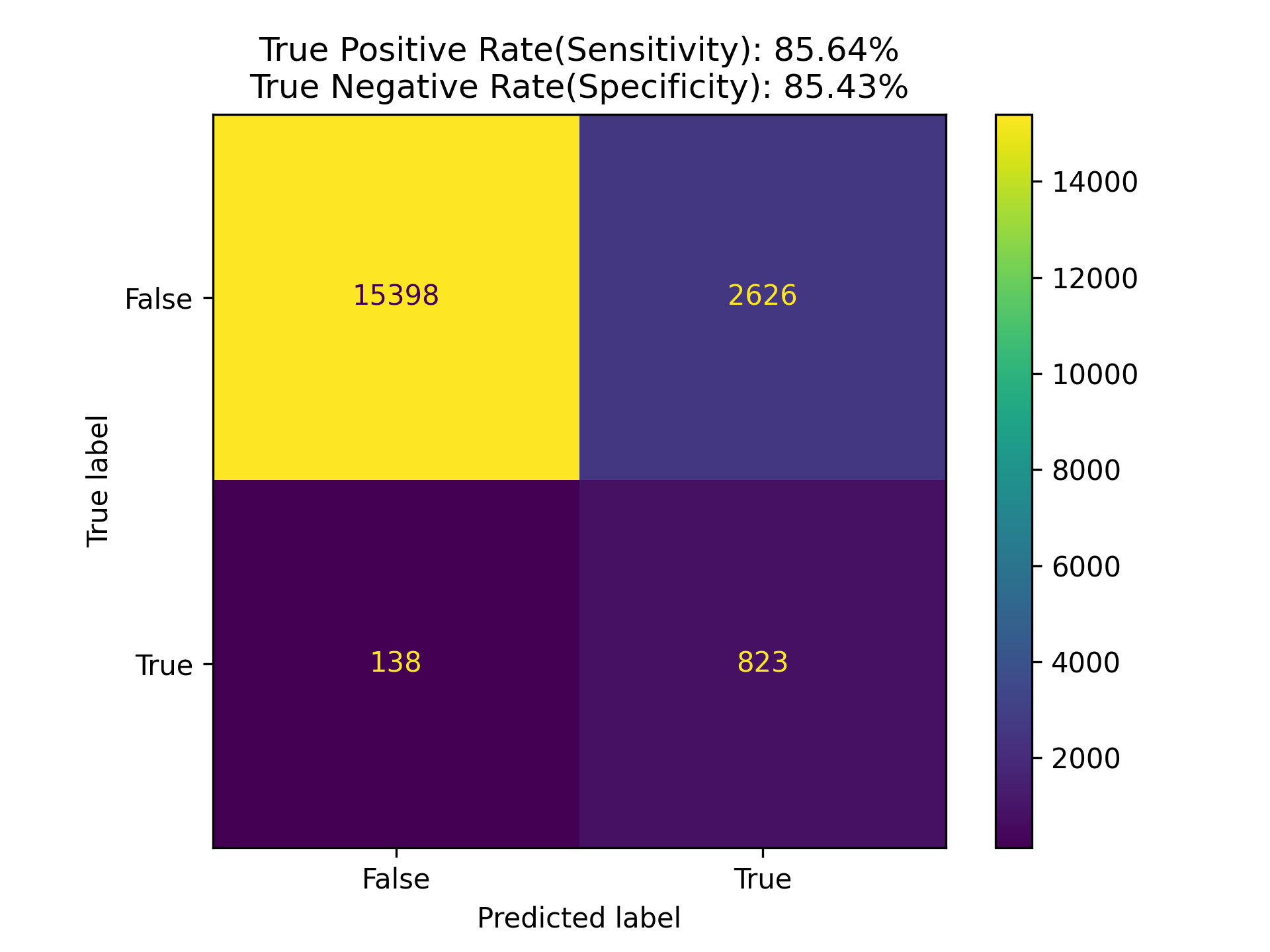}
            \begin{center}
            (a)
            \end{center}
        \end{minipage}\hfill
        \begin{minipage}{0.5\columnwidth}
            \centering
            \includegraphics[width=\columnwidth]{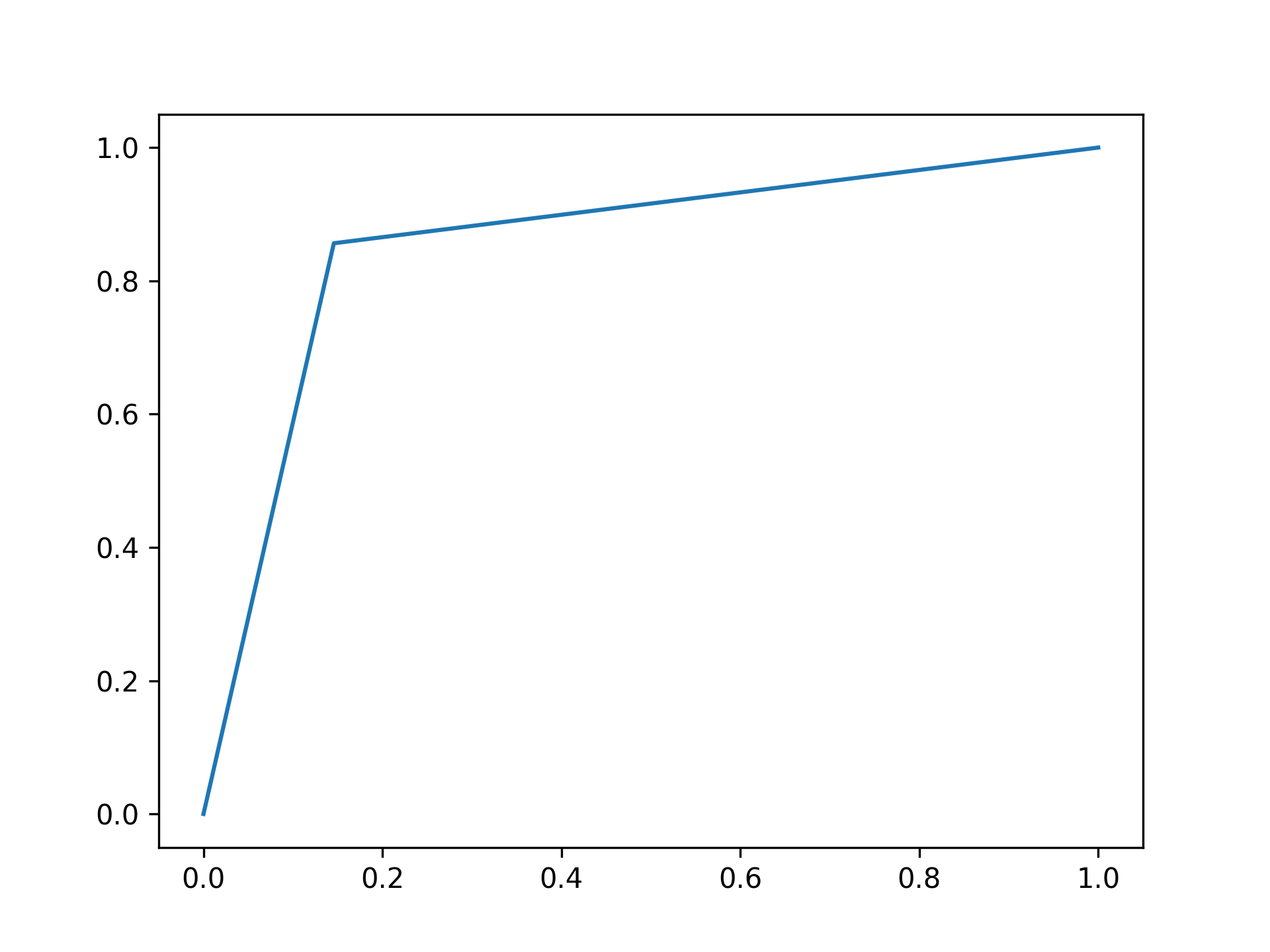}
             \begin{center}
            (b)
            \end{center}
        \end{minipage}
        \caption{(a) Confusion matrix (left) and (b) ROC curve (right) for hybrid model results}
        \label{fig:hybrid_conf_results}
\end{figure}

As we can see from Figure \ref{fig:combined_epoch_loss}, the loss successfully
converged and reached in a range of $0.07$ to $0.08$.
Our top priority, $TPR$ reaches $85\%$ while the $TNR$ also
reached $85\%$ during that within that range of losses.
Therefore, our accuracy is also around $85\%$ as calculated by $\frac{TPR + TNR}{P+N}$,
as shown in Figure \ref{fig:hybrid_conf_results}.
However, we can see from both of the confusion matrices in Figure \ref{fig:confusion_ABC} (c)
and Figure \ref{fig:hybrid_conf_results}, the precisions are relatively
low since inside the test case for undersampling, we move some majorities to the test case.
This is one of the drawbacks of this model.
If we have a low precision, it means
that there is a high probability that there would not be a fire,
but we categorized it as true, which causes a false alarm, and a type II error will occur.

Usually, based on the experiences, during training, about $60$ epoch,
the $TPR$ and $TNR$ will both reach around $80\%$ and so does accuracy.
When reaching $100$ epochs, all of them might be more than $83\%$.

\subsection{Main results}

From Table \ref{tab:results_summary}, we summarized the results of our experiments.
As shown, the best $TNR$ is in all the features without images and undersampling method.
Even though with a $TPR$ about $70\%$, the precision is relatively high, which means
that the model can predict accurately within the true cases.
However, there is a trade-off, with the $TPR$ increase, our $TNR$ experiment
results go down $10\%$,
as shown in Table \ref{tab:results_summary}.
So does the precision, since the precision is related to the proportion
of our data as shown in Appendix \ref{apd:precision}, among the $18955$ test cases,
there are only $5\%$ with the true label,
even $85\%$ of them are categorized right, the precision, $\mathbb{P}(actual = True|predict = True)$,
went down to about $23.86\%$ from Figure \ref{fig:hybrid_conf_results},
 with a $5\%$ increase from the results, $18.78\%$,
of the features without images.

\begin{table}[!htb]
    \centering
    \begin{tabular}{|c|c|c|c|c|c|}
        \hline
        Model & Features Used & loss & Top Accuracy(\%) & Top TPR(\%) & Top TNR(\%)\\
        \hline
        \hline
        \multirow{4}{*}{Baseline} & t,w & $\geq 0.12$ & $\leq 85$ & $40 \sim 50$ & $\approx 82$ \\
        \cline{2-6}
        & t,w,h,p & $\geq 0.06$ & $88 \sim 89$ & $63 \sim 68$ & $92 \sim 95$ \\
        \cline{2-6}
        & t,w,h,p,v & $\approx 0.06$ & $89 \sim 90$ & $65 \sim 70$ & $\approx 95$ \\
        \cline{2-6}
        & undersample all & $\approx 0.06$ & $\geq 80$ & $75 \sim 82 $ & $ 80 \sim 81 $ \\
        \hline
        % \multirow{2}{*}{Combined Model} & image & & & & \\
        % \cline{2-6}
        Hybrid & undersample image & $\leq 0.1$ & $\approx 85$ & $ \approx 85 $ & $ \approx 85 $\\
        \hline
    \end{tabular}
    \caption{top $5\%$ of metrics, note:t=temperature, w=wind, h=humidity, p=precipitation, v=vegetation, all=all features}
    \label{tab:results_summary}
\end{table}

\begin{figure}[!htb]
     \centering
     \begin{minipage}{0.5\textwidth}
         \centering
         \includegraphics[width=\textwidth]{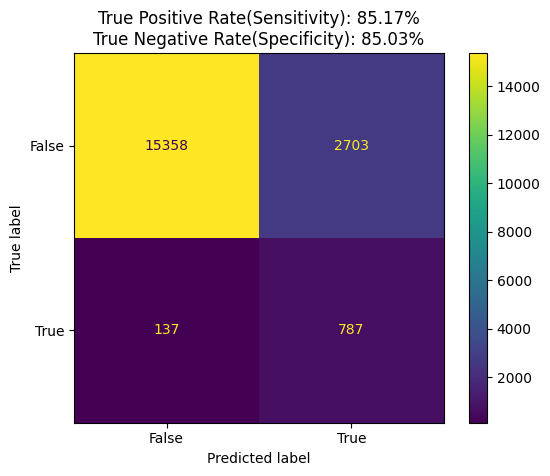}
         \begin{center}
         (a)
         \end{center}
     \end{minipage}\hfill
     \begin{minipage}{0.5\textwidth}
         \centering
         \includegraphics[width=\textwidth]{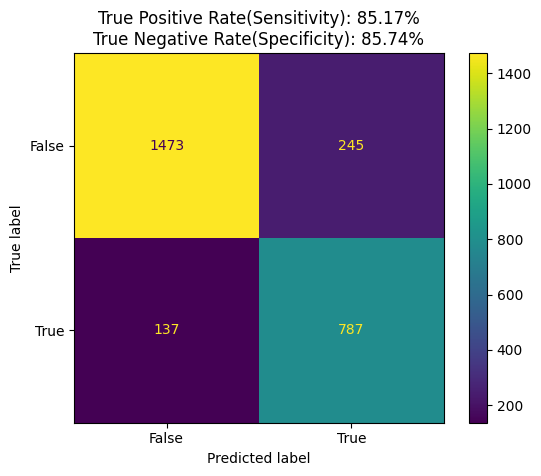}
         \begin{center}
         (b)
         \end{center}
     \end{minipage}
     \caption{
     (a) Confusion matrix with full test, and (b)Confusion matrix without majorities.
     }
     \label{fig:Confusion}
\end{figure}

As we can see from Figure \ref{fig:Confusion} (b), another optimal case,
the results without adding the majority to the test case, we still have about $85\%$ accuracy.
Since all true cases are in the original test cases the $TPR$ will not change.
We can see that from both Figures \ref{fig:Confusion} (a) and \ref{fig:Confusion} (b),
for the added majorities, which has $16343$ cases with all
false, the TN is about $13885$ and FP is $2458$, Thus, the $TNR$ is
about $84.96\%$ which is also the accuracy in this case since there are no true results in this part.
However, if we do not include  the majority of test cases, the precision
can reach about $\frac{787}{245+787}=76.26\%$, since our overall accuracy is still
about $85\%$, with more false cases, the precision will go down.
Therefore, the proportion of the data will have a large influence on the precision.
If we include more true data later, the precision should go up.
Moreover, once both $TPR$ and $TNR$ are improved, the precision should also be improved.

\subsection{Case Study: 2018 California Camp Fire}
\begin{figure}[!htb]
    \centering
    \includegraphics[width=.85\columnwidth]{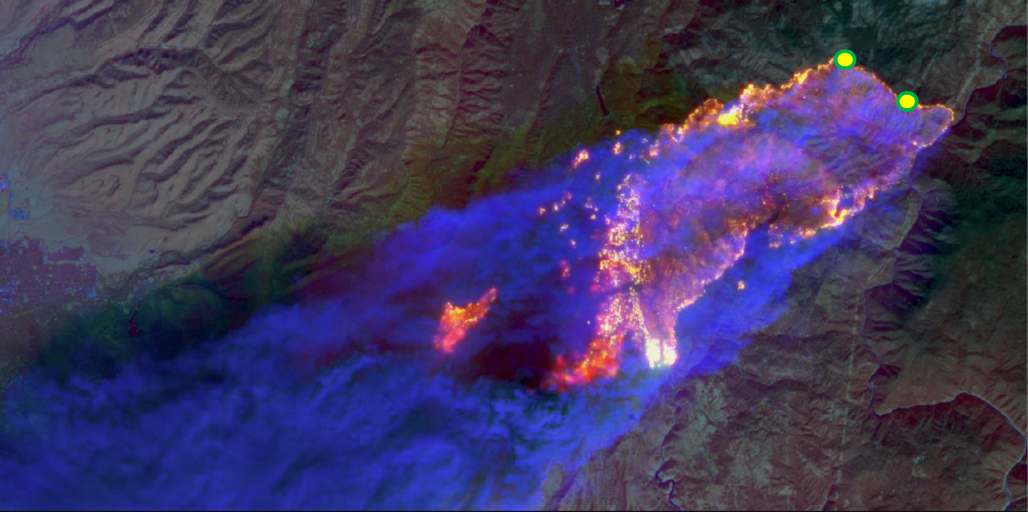}
    \caption{Landsat 8 overpass of the Camp Fire on November 8, 2018,
     at 1:44 pm EST (10:44 am PST).
     Credits:NASA/Marshall Space Flight Center/United States Geological Survey
     \cite{Sheaves2021})}
    \label{fig:Pulga-CA1}
\end{figure}

To demonstrate and validate the ability and effectiveness of
the proposed high-resolution multimodal transformer neural network model
to forecast or predict wildfire occurrence,
we used the local weather forecasting data on November 7th, 2018,
and the local Google Earth image data before November 7th, 2018, in
the vicinity of Pulga, California,
just right before the occurrence of the 2018 California Camp (November 8th, 2018)
to test the performance of the proposed Bayesian machine learning model.

Figure \ref{fig:Pulga-CA1} is a real satellite image taken right after
the starting occurrence of the California Camp Fire
near the town of Pugla, California, on November 8th, 2018, by NASA.

The two fire staring or occurrence locations are marked as colored circles (I: Orange circle
and II: Red circle.) in Figure \ref{fig:Pulga-CA1}.

\begin{figure}[!htb]
    \centering
    \includegraphics[width=.85\columnwidth]{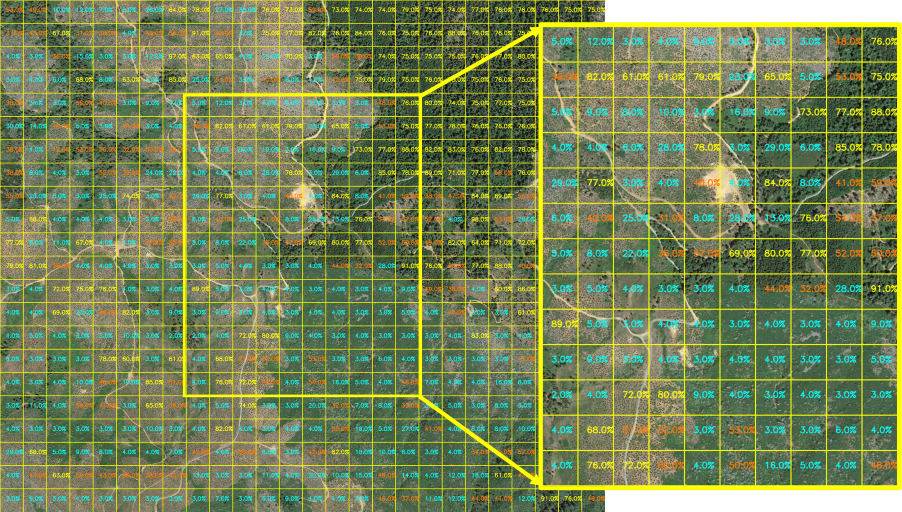}
    \caption{Prediction results for the 2018 California Camp Fire occurrence
    location I.}
    \label{fig:Pulga-CA2}
\end{figure}

By inputting the local weather data and the image data from Google Earth into
the multimodal transform neural network, the multimodal transformer machine
learning neural network
will output the probability distribution in the local region.
Figure \ref{fig:Pulga-CA2} is the prediction result obtained
by using the proposed multimodal transformer neural network
model on prediction of the occurrence of the California Camp Fire
at the observed fire occurrence site I.
The topographic and weather information input data and parameters are given as follows,
\\
\noindent
[39.831119, -121.480039,
18.74354839,   18.240000,   18.30714286,
\\
16.20967742,      16.680000,    16.57142857,
  37.60000,      39.93333333,  \\
      34.61428571,
0.0000,       0.0000,     0.0000]\\
where the first two are the converted longitude and latitude of the location.
The next 12 parameters are the average temperature, wind speed, and precipitation,
humidity for the prior 7, 15, and 30 days before November 8th, 2018.
The last 6 are the vegetation categories for this location.
This row is repeated 195 times since this area shares the same data value,
the only difference is the 195 different input images as shown in Figure \ref{fig:Pulga-CA2}.

The predicted wildfire probabilities on each small-scale box
of $100m \times 100m$ are depicted in Figure \ref{fig:Pulga-CA2}.
We marked all the wildfire occurrences with high probability ($> 70\%$) regions
with yellow color probability numbers.

For the fire occurrence location II,
we also input the local geographic and weather data, as well as
Google Earth image data into the Bayesian machine learning
neural network. The input information data is given as follows,

\noindent
[ 39.825097,-121.446211,
18.74354839,   18.240000,   18.30714286,
\\
 16.20967742,      16.680000,    16.57142857,
 37.60000,      39.93333333,
\\
34.61428571,
0.0000,       0.0000,     0.0000]
\\
\smallskip

\begin{figure}[!htb]
    \centering
    \includegraphics[width=.85\columnwidth]{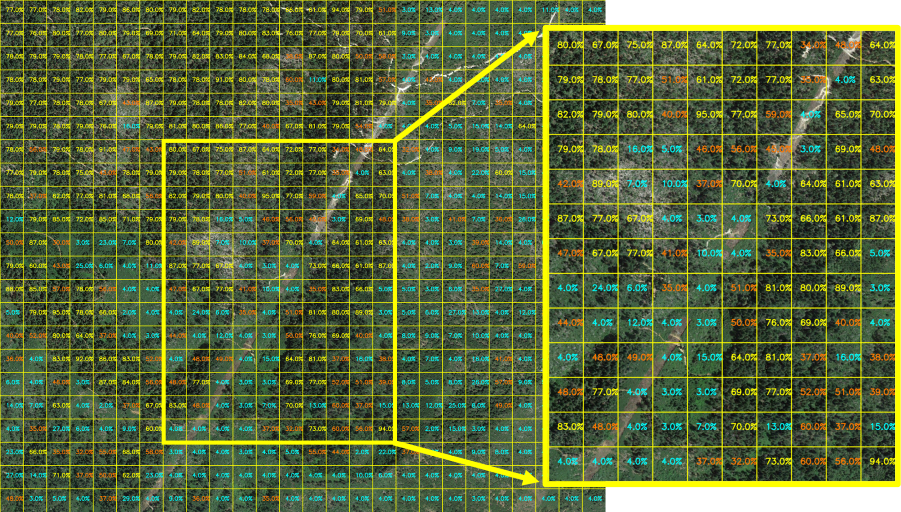}
    \caption{Prediction results for the 2018 California Camp Fire occurrence
    location II.}
    \label{fig:Pulga-CA3}
\end{figure}

The predicted wildfire probabilities on each small-scale box
of $100m \times 100m$ are depicted in Figure \ref{fig:Pulga-CA3}.
Again, we marked all the wildfire occurrences with high probability ($> 70\%$) regions
with yellow color probability numbers as shown in Figure \ref{fig:Pulga-CA3}.

It is worth noting that the training data that we used is from 1992 to 2015,
while the California Camp Fire occurred in 2018.
Therefore, this example is a bona fide forecasting demonstration.

From Figures \ref{fig:Pulga-CA2} and \ref{fig:Pulga-CA3},
One may find that the places where wildfires most likely occur
are regions with dense vegetation and flat terrain.

\section{Conclusions and Perspectives}

In this work, we developed a high-resolution multimodal transformer neural network
model to forecast wildfire hazard occurrence location and timing at a small scale.
Different from the existing ML models for wildfire prediction, the present
approach combines weather information, historical fire data,
local topographic information, and Google Earth image information to train
the multimodal transformer neural
network machine learning model,
and then practically provides the wildfire occurrence forecast
at a local spatial location within $\le 100 m^2$.
By using the wildfire data in the United States from 1992 to 2015 (USDA\cite{fpa_fod2017})
to train the high-resolution multimodal transformer neural network,
we can provide the small-scale spatial probability of wildfire occurrence
with the input of the given weather forecasting information and
the synchronized local Google Earth imagery information 24 hours before.
This is a significant advance in wildfire forecasting and prediction technology, and we have made a major
step forward in wildfire occurrence forecasting.

Moreover, in this work, we use the wildfire data collected in
the $FPA\_FOD$ database to demonstrate the viability and the
accuracy of the proposed artificial intelligence-based wildfire
forecast system.
The fourth edition of $FPA\_FOD$ ([7]) contains 1.88 million entries of wildfire records from 1992 to 2015.
Recently, a new version with 2.3 million entries of wildfire records was released to the public \cite{fpa_fod2022}.
It contains a spatial database of wildfires that occurred in the US
from 1992 to 2022, which is the 7th edition of the wildfire database
to support the National Fire Program Analysis (FPA) system\cite{fpa_fod2022}.

Thus, based on this new database, we can use a database with more data information,
and more data features,
and better-quality data to train the proposed Bayesian machine learning
neural network.
The new database version increases about $42$ thousand wildfire records within $5$ years.
As mentioned before, some of the images from Google Earth
are in grayscale before $2005$,
however, with this newly released data,
we should be able to use the colored images of vegetation
and extract terrain topological features.

Moreover, all the occurrences from $FPA\_FOD$ are within the US, and recently, a data called \href{https://www.kaggle.com/datasets/abdelghaniaaba/wildfire-prediction-dataset}
{Wildfire Prediction Dataset (Satellite Images)} was also released\cite{aaba2023}.
It is based on wildfire data from Canada's Open Government Portal\cite{canada_fire}.
It contains two categories: wildfire images and no wildfire images. We can use wildfire images to increase and enrich the data set.
Furthermore, we may even be able to add more data, such as Global Fire Emissions Database version 4\cite{gfedv4}.
Combining with and extending remote sensing data\cite{SAYAD2019130}
that was collected from MODIS (Moderate Resolution Imaging Spectroradiometer).
Besides the data collected from North America.

Moreover, all the occurrences from $FPA\_FOD$ are within the US, and recently, a data called \href{https://www.kaggle.com/datasets/abdelghaniaaba/wildfire-prediction-dataset}
{Wildfire Prediction Dataset (Satellite Images)} was also released\cite{aaba2023}.
It is based on wildfire data from Canada's Open Government Portal\cite{canada_fire}.
It contains two categories: wildfire images and no wildfire images. We can use wildfire images to increase and enrich the data set.
Furthermore, we may even be able to add more data, such as Global Fire Emissions Database version 4\cite{gfedv4}.
Combining with and extending remote sensing data\cite{SAYAD2019130}
that was collected from MODIS (Moderate Resolution Imaging Spectroradiometer).

Besides the data collected from North America. the wildfire data in
other parts of the world may also be available \cite{RODRIGUEZASERETTO20131861}.
With more data, not only the neural network model could perform better,
but the transformer like the model, will also have a better performance
since, as we mentioned before, a transformer usually requires a large amount of data\cite{9944625}.
As mentioned in \ref{limitations},
the accuracy of the data may significantly affect the prediction outcome.
Therefore, for a certain location, if the data quality could be better,
especially for the image data, the accuracy could significantly be increased.
In the present work,
some of the features that are still missing from the database,
such as elevation or sensing data.
Remote sensing data can also be added to the data if the amount
data collected is enough,
which can also help predict wildfire occurrence\cite{rs12142246}.
We shall report an updated version of the multimodal transformer
neural network model for wildfire prediction in subsequent work.

\vskip 1.0in

% \section*{References}

\section*{References}
\bibliographystyle{unsrt}
\bibliography{references}
% \bibliographystyle{plain}
% \bibliography{references}

\newpage

\appendix
\nolinenumbers

%\linenumbers
\pagestyle{plain}
\setcounter{page}{1}
\centerline{\large \bf Supplementary Materials}

\section*{1. README File}

In this Supplementary material,
we provide a README file for readers to assist anyone who wishes
to run the related computer code
of the high-resolution multimodal transformer neural network model described in this paper.
The interested readers can write to the corresponding author of this paper requesting
the source code of the computer program. The reader should be
able to run this code either on your own machine or on Google Colab.
\begin{enumerate}
    \item Local machine: If running on local machine, please run
        \begin{lstlisting}[language=Bash,]
            $ pip install -r requirements
        \end{lstlisting} or during running it will automatically checking satisfaction for requirements
    \item Online Colab: First cell will check the requirements. Colab is available \href{https://www.tinyurl.com/2p9ezkzm}{here}
\end{enumerate}

\label{colab_instruct}

\subsection*{1.1 Requirements}

\begin{itemize}
    \item Python 3.6+
    \item PyTorch 2.0.0+\cite{NEURIPS2019_9015}
\end{itemize}

\subsection*{1.2 Usage}

The script and data are available from
\begin{table}[!htb]
    \centering
    \begin{tabular}{|c|c|c|}
        \hline
        Data Name & Size \\
        \hline
        \hline
        \href{https://drive.google.com/uc?export=download&id=1iCYPGnBsvbw7gpFmF4w_ag29vDx3MZS9}{Scripts \& Codes} & $47K$ \\
        \hline
        \href{https://drive.google.com/uc?export=download&id=11lKss_T2m3rzLkS3XYK6aa3fkrxwdYmN}{Information Data} & $6.29M$\\
        \hline
        \href{https://drive.google.com/uc?export=download&id=1oRvPqGTg9sIWgiEmIyboCzb31g-SV-nW}{Imagery Data} & $1.10G$\\
        \hline
        Original Image Captures & TBD\\
        \hline
        \href{https://drive.google.com/uc?export=download&id=1MqF5qWbjmL6ZjTuov4smHWeYgz5tU6pA}{Pre-trained model} & $34.2M$ \\
        \hline
        \href{https://www.fs.usda.gov/rds/archive/products/RDS-2013-0009.4/_metadata_RDS-2013-0009.4.html}{FPA\_FOD\_20170508 Wildfire information} & NA\\
        \hline
        \href{https://www.ncei.noaa.gov/pub/data/noaa/}{NOAA weather information} & NA \\
        \hline
    \end{tabular}
    \caption{Data URL}
    \label{tab:dataset_summary}
\end{table}

\medskip

\subsubsection*{1.2.1 Run the code locally}
\label{apd:local_cmd}
Example command for run a combined model:
\begin{lstlisting}[language=Bash,]
    $ python train.py --with-vegetation=True\
                    --satellite-img=True\
                    --gray-scale=True\
                    --resample-method='undersampling'\
                    --archive=True\
                    --model-selection='hybrid model'\
                    --learning-rate=0.01\
                    --epochs=100\
                    --display_step=10\
                    --batch=True\
                    --batch-size=64

\end{lstlisting}
This command in terminal will allow training the model vegetation and image features
using undersampling method.
The model is hybrid model with 64 mini-batchs.
It will run 100 epochs with learning rate 0.01, for every 10 steps, and then
it will display the loss or accuracy.
After finishing training, the model will be saved to './saved\_progress'.

You can also run it with
\begin{lstlisting}[language=Bash,]
    $ python train.py - h
\end{lstlisting}
to see help and all available arguments.

Or using UI as following by running
\begin{lstlisting}[language=Bash]
    $ python wildfire_ui.py
\end{lstlisting}
\begin{figure}[!htb]
    \centering
    \includegraphics[width=.5\columnwidth]{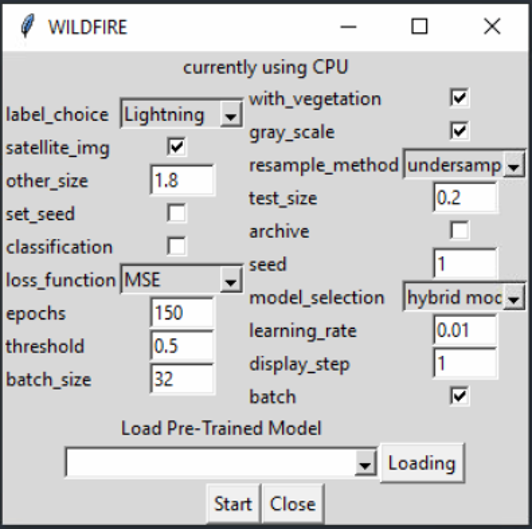}
    \caption{User Interface}
    \label{fig:ui}
\end{figure}

This UI will not run the program,
instead it will generate a command state in \ref{apd:local_cmd}.

\medskip

\subsubsection*{1.2.2 Instructions on Colab}

First time run steps:
\begin{enumerate}
    \item Loading Required Package
    \item Create Symbolic Link
    \item Util Functions
    \item Parameter Setting
    \item Read Subset Data \& Create DataFrame
    \item Create Dataset (Train set \& Test set)
    \item Neural Network (or F-CNN \& ResNet)*
    \item Hybrid Model
    \item Initialize model
    \item Training
\end{enumerate}

\begin{itemize}
    \item Repeat training:
    \begin{itemize}
        \item Training (step 10)
    \end{itemize}
    \item Re-train a new model with same data
    \begin{itemize}
        \item Initialize model (step 9)
        \item Training (step 10)
    \end{itemize}
    \item Train a new model with new shuffle data
    \begin{itemize}
        \item Create Dataset (Train set \& Test set) (step 6)
        \item Initialize model (step 9)
        \item Training (step 10)
    \end{itemize}
    \item Train a new model with some arguments change
    \begin{itemize}
        \item Parameter Setting (step 4)
        \item Read Subset Data \& Create DataFrame (step 5)**
        \item Create Dataset (Train set \& Test set) (step 6)**
        \item Initialize model (step 9)
        \item Training (step 10)
    \end{itemize}
    \item Results
    \begin{enumerate}
        \item Accuracy
        \item Confusion matrix
        \item Tensorboard
    \end{enumerate}
    \item Load pre-trained model***
    \begin{lstlisting}[language=Python,]
        import torch
        model = torch.load(<desired model path>)
    \end{lstlisting}
\end{itemize}
Here the symbol {*} indicates it applying to any desired models
\\
\noindent
The symbol {**} indicates it applying when change input data (i.e test size, undersampling)
\\
\noindent
The symbol {***} indicates that the saved progress may include 8 files\\

\begin{tikzpicture}
    [level 1/.style = {blue},
    every node/.append style = {draw, anchor = west},
    grow via three points={one child at (0.5,-0.8) and two children at (0.5,-0.8) and (0.5,-1.6)},
    edge from parent path={(\tikzparentnode\tikzparentanchor) |- (\tikzchildnode\tikzchildanchor)}]

    \node {folder path}
        child {node {1 event file}}
        child {node {1 model file}}
        child {node {1 pt file}}
        child {node {2 loss pngs}}
        child {node {1 confusion matrix png}}
        child {node {1 roc curve png}}
        child {node {1 index file}};
\end{tikzpicture}
\\

Event file stores the log information for training and test,
model and .pt files store the model and all parameters separately,
loss PNGs are the plots for the loss based on epochs or batch steps,
confusion matrix png file displays the confusion matrix as a table,
roc curve png show the ROC curve for the confusion matrix, and index
file contains the random indices when data was shuffled and split
for training and test initially for reuse purpose.
\\

\begin{table}[!htb]
    \centering
    \begin{tabular}{c|c|c}
    \hline
    Params & Baseline & Hybrid \\
    \hline
    w/\ veg & True & True\\
    \hline
    udersample & $ \geq 1.5 $ & $ \geq 1.5 $\\
    \hline
    test size & $\geq 0.15$ & $ \geq 0.15 $\\
    \hline
    learning rate & 0.01 & 0.01 \\
    \hline
    epochs & $\leq 300$ & $ \leq 50 $\\
    \hline
    \end{tabular}
    \caption{Recommended Tuning}
    \label{tab:recomd_tune}
\end{table}

\medskip

\newpage
\subsection*{1.3 Parameter Descriptions}

\begin{table}[h]
    \centering
    \begin{tabular}{p{0.25\linewidth} | p{0.4\linewidth} | p{0.15\linewidth}} %{c|c|c}
        \hline
        Parameters & Descriptions & Type \\
        \hline
        \hline
        connect to google drive & allowing script connect to user's Google Drive & boolean \\
        \hline
        download\_pre\_trained & download pre-trained model & boolean \\
        \hline
        label choice & choose desired label & string \\
        \hline
        with vegetation & check box for vegetation categories as features & boolean \\
        \hline
        satellite images & check box for satellite images as features & boolean \\
        \hline
        resample method & choose resample methods & string \\
        \hline
        other size &  with previous checked, choosing proportion for undersampling & float \\
        \hline
        test size & the proportion of test set & float \\
        \hline
        archive & check box for save the results & boolean \\
        \hline
        set seed & check box for control the randomness & boolean \\
        \hline

        seed & the seed for randomness & integer \\
        \hline
        model selection & select training models & string \\
        \hline
        loss function & choose loss function & string \\
        \hline
        learning rate & learning rate for optimizer & float \\
        \hline
        epochs & the amount of iterations training will run & integer \\
        \hline
        display steps & show the loss and accuracy & integer \\
        \hline
        threshold & a number larger than it will be false otherwise true & float \\
        \hline
        batch & check box for allowing batch during training & boolean \\
        \hline
        batch size & with previous checked, the size of each batch & integer \\
        \hline
    \end{tabular}
    \caption{Parameter Settings Explanations}
    \label{tab:params_explan}
\end{table}
\label{apd:recomm_tune}

\bigskip
\bigskip

\section*{2. Data Content Description}
\label{data_desc}

In this Section, we provide the information data input format of this study,
which is used as the data input format.
They are given as follows.
\begin{table}[!htb]
    \centering
    \begin{tabular}{p{0.4\linewidth} | p{0.5\linewidth}}
        \hline
        FOD$\_$ID & Global unique identifier\\
        \hline
        \hline
        FIRE$\_$NAME & Name of the incident, from the fire report (primary) or ICS-209 report (secondary).\\
        \hline
        date(julianday(DISCOVERY$\_$DATE)) & Date on which the fire was discovered or confirmed to exist\\
        \hline
        DISCOVERY$\_$TIME & Time of day that the fire was discovered or confirmed to exist\\
        \hline
        STAT$\_$CAUSE$\_$CODE & Code for the (statistical) cause of the fire\\
        \hline
        STAT$\_$CAUSE$\_$DESCR & Description of the (statistical) cause of the fire\\
        \hline
        LATITUDE & Latitude (NAD83) for point location of the fire (decimal degrees)\\
        \hline
        LONGITUDE & Longitude (NAD83) for point location of the fire (decimal degrees).\\
        \hline
        FIRE$\_$SIZE$\_$CLASS & Code for fire size based on the number of acres within the final fire perimeter expenditures\\
        \hline
        FIRE$\_$SIZE & Estimate of acres within the final perimeter of the fire\\
        \hline
    \end{tabular}
    \caption{FPA\_FOD description\cite{fpa_fod2017}}
    \label{tab:fpa_fod_param_desc}
\end{table}
		
\bigskip
\bigskip

\section*{3. General Wildfire Causes defined in National Wildfire Coordinating Group}
\label{nwcg_causes}
In this Section, we list all known or defined causes of a wildfire
in National Wildfire Coordinating Group (NWCG).
\begin{table}[!htb]
    \centering
    \caption{Gerneral Causes by National Wildfire Coordinating Group\cite{nwcg}}
    \begin{tabular}{|c|c|}
        \hline
        Existing NASF Cause Code & Proposed Fire Cause Standard \\
        \hline
        1-Lightning  & Natural \\
        \hline
        2-Equipment Use & Equipment and vehicle use \\
        \hline
        3-Smoking & Smoking \\
        \hline
        4-Campfire & Recreation and ceremony \\
        \hline
        5-Debris Burning & Debris and open burning \\
        \hline
        6-Railroad & Railroad operations and maintenance \\
        \hline
        7-Arson & Arson \\
        \hline
        8-Children & Misuse of fire by a minor \\
        \hline
        9-Miscellaneous & Other causes \\
        \hline
        10-Fireworks & Fireworks \\
        \hline
        11- Power line & Power generation/transmission/distribution \\
        \hline
        12-Structure & \sout{Structure} (under Other)  \\
        \hline
        & Firearms and explosive use \\
        \hline
        & Undetermined (Human or Natural) \\
        \hline
    \end{tabular}
    \label{tab:nwcg}
\end{table}

\bigskip

\newpage
\section*{4. Vegetation Categories}
In this Section, we provide a list of vegetation categories that
are used in the present study,
which is given as follow.

\begin{table}[h]
    \centering
    \begin{tabular}{c|c}
        \hline
        Vegetation & Dominant vegetation in the areas \\
        \hline
        1 & Tropical Evergreen Broadleaf Forest \\
        \hline
        2 & Tropical Deciduous Broadleaf Forest \\
        \hline
        3 & Temperate Evergreen Broadleaf Forest \\
        \hline
        4 & Temperate Evergreen Needleleaf Forest TmpENF \\
        \hline
        5 & Temperate Deciduous Broadleaf Forest \\
        \hline
        6 & Boreal Evergreen Needleleaf Forest \\
        \hline
        7 & Boreal Deciduous Needleleaf Forest \\
        \hline
        8 & Savanna \\
        \hline
        9 & C3 Grassland/Steppe \\
        \hline
        10 & C4 Grassland/Steppe \\
        \hline
        11 & Dense Shrubland \\
        \hline
        12 & Open Shrubland \\
        \hline
        13 & Tundra Tundra \\
        \hline
        14 & Desert \\
        \hline
        15 & Polar Desert/Rock/Ice \\
        \hline
        16 & Secondary Tropical Evergreen Broadleaf Forest \\
        \hline
        17 & Secondary Tropical Deciduous Broadleaf Forest \\
        \hline
        18 & Secondary Temperate Evergreen Broadleaf Forest \\
        \hline
        19 & Secondary Temperate Evergreen Needleleaf Forest \\
        \hline
        20 & Secondary Temperate Deciduous Broadleaf Forest \\
        \hline
        21 & Secondary Boreal Evergreen Needleleaf Forest \\
        \hline
        22 & Secondary Boreal Deciduous Needleleaf Forest \\
        \hline
        23 & Water/Rivers Water \\
        \hline
        24 & C3 Cropland \\
        \hline
        25 & C4 Cropland \\
        \hline
        26 & C3 Pastureland \\
        \hline
        27 & C4 Pastureland, 28:Urban land\\
        \hline
    \end{tabular}
    \caption{Vegetation Categories}
    \label{tab:veg_cat}
\end{table}

\bigskip
\bigskip

\section*{5. Synthetic Minority Over-sampling Technique(SMOTE)}
\label{apd:smote}

In this Section, we briefly outline the Synthetic Minority Over-sampling Algorithm
as follows.
\begin{algorithm}[H]
    \caption{Synthetic Minority Over-sampling Technique (SMOTE)\cite{Chawla_2002}}
    \begin{algorithmic}
        \State Calculate Euclidean Distance between every samples with minorities $S_{min}$ to find $k-nearest$
        \State Find $N$, based on proportion of minorities and majorities,
        for each minority sample $X$, randomly choosing from its nearest sample $X_{n}$
        \FOR {each $X_{n}$}
            $X_{new} = X + random(0,1) \times |X-X_{n}|$
        \ENDFOR
    \end{algorithmic}
\end{algorithm}

\bigskip

\section*{6. Precision}
\label{apd:precision}

In this Section, based on the conditional probability of a random event,
we provide a mathematical formula
to calculate the precision of the developed high-resolution
Bayesian learning neural network model, which is
given as follows.
\begin{align*}
    Precision &= \mathbb{P}(actual = 1 | predict =1)\\
    &= \frac{\mathbb{P}(actual =1, predict = 1)}{\mathbb{P}(predict =1)}\\
    &= \frac{\mathbb{P}(predict = 1 | actual = 1) \mathbb{P}(actual = 1))}{\mathbb{P}(predict = 1, actual = 1) + \mathbb{P}(predict = 1, actual = 0)}\\
    &= \frac{\mathbb{P}(predict = 1 | actual = 1) \mathbb{P}(actual = 1))}{\mathbb{P}(predict = 1 | actual = 1) \mathbb{P}(actual = 1)) + \mathbb{P}(predict = 1 | actual = 0) \mathbb{P}(actual = 0))}\\
    & = \frac{TPR \cdot n}{TPR \cdot n + FPR \cdot (1-n)}\\
\end{align*}
where n is the proportion of the actual true.
Based on the equation above, precision is related to the proportion of true and false data,
once the $\frac{FPR}{TPR}$ is fixed, it increases with $n$ increases.
However, TPR and FPR are not related to the proportion of the data,
since $TPR = \mathbb{P}(predict = 1 | actual = 1)$ and $FPR = \mathbb{P}(predict = 1 | actual = 0)$\cite{10.1145/1143844.1143874}.

\bigskip

\section*{7. Example of Sample Outputs}
\label{apd:sample_out}
\begin{figure}[!htb]
    \centering
    \text{\#1333}\\
    \includegraphics[width=3.0in]{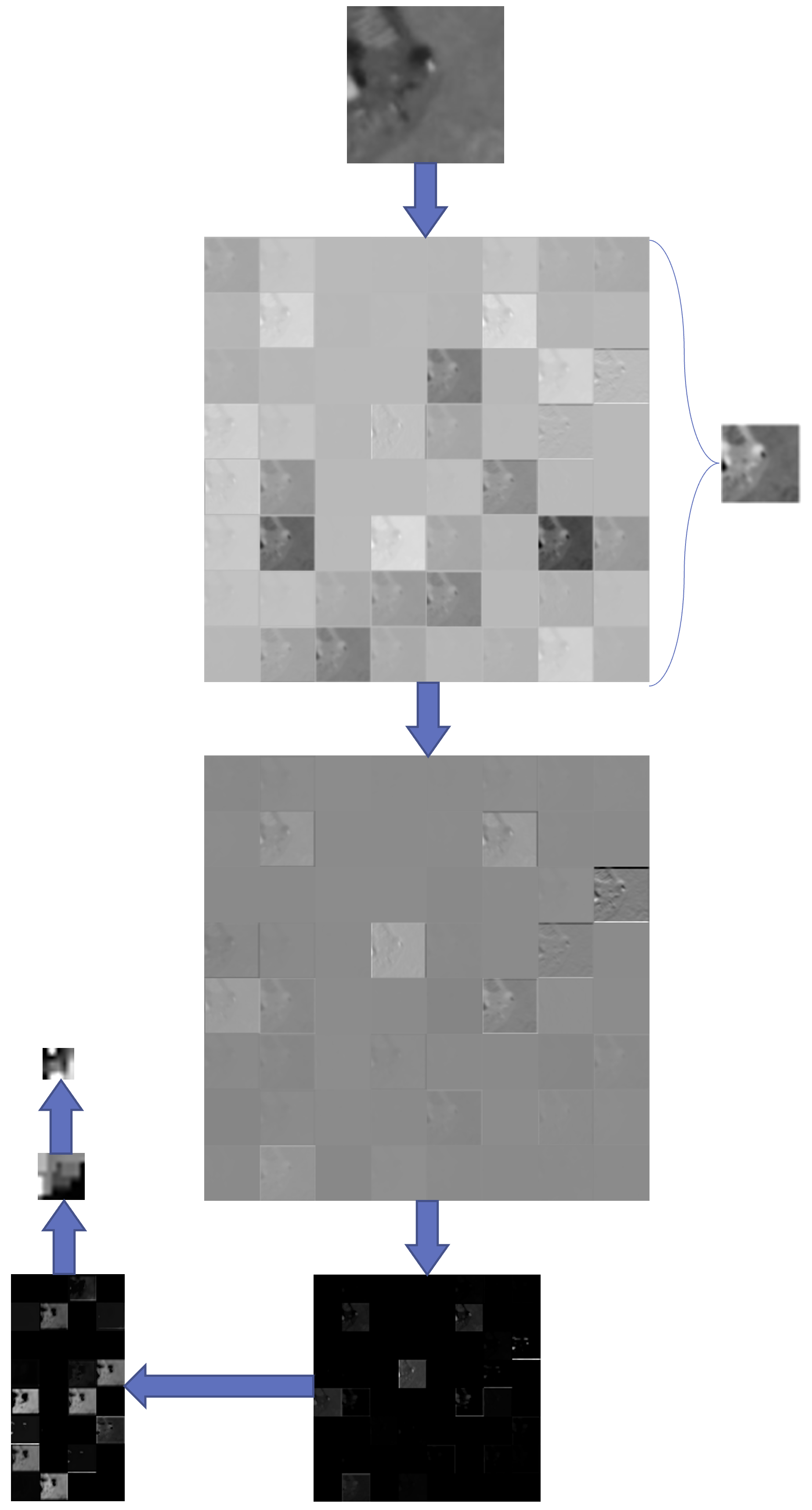}
    \caption{Image flow of image \#1333 for each layer. From top: original
    input $\rightarrow$ 1st
    convolution output $\rightarrow$ 1st batch normalization output $\rightarrow$
    1st maxpooling output $\rightarrow$ 1st residual block output $\rightarrow$ WIT output}
    \label{figure:sample_out}
\end{figure}

In this Section, we show an example of sample outputs.
Figure \ref{figure:sample_out} displays the sample outputs
of the image No. 1333.

\end{document}